\documentclass[3p,12pt]{elsarticle}

\usepackage{amsmath,amssymb,bm}

\usepackage{graphicx}
\usepackage{subcaption}
\usepackage{float}
\graphicspath{{../}}             

\usepackage{booktabs}
\usepackage{multirow}
\usepackage{tabularx}

\usepackage[ruled,vlined,linesnumbered]{algorithm2e}

\usepackage{enumitem}
\usepackage{url}
\usepackage{hyperref}

\journal{IJMR}

\begin{document}

\begin{frontmatter}

\title{Vision-Guided Dual-Arm Humanoid Robotic Disassembly of End-of-Life 18650 Lithium-ion Battery Packs}

\author[kth]{Yile Chen}

\author[kth]{Zhihao Liu\corref{cor1}}
\ead{zhihaoliu@ieee.org}
\cortext[cor1]{Corresponding author.}

\author[kth]{Xi Vincent Wang}
\author[kth]{Lihui Wang}

\affiliation[kth]{organization={KTH Royal Institute of Technology},
  addressline={Brinellv\"agen 8},
  postcode={114 28},
  city={Stockholm},
  country={Sweden}}

\begin{abstract}
The growing volume of retired lithium-ion battery packs from
electric vehicles and portable electronics calls for automated
disassembly that is safe, flexible, and selective down to the
individual cell. Existing robotic systems, however, mostly assume
known pack poses, external fixtures, or specialised tooling,
leaving fixture-free cell-level disassembly under pose uncertainty
largely unsolved. This paper presents a vision-guided dual-arm
pipeline that disassembles a 21-cell 18650 pack from an arbitrary
initial pose using only general-purpose parallel-jaw grippers,
RGB-D sensing, and a pre-trained grasp detector. Pose uncertainty is
absorbed by a learn-and-filter perception stack with discrete
look-and-move wrist-camera corrections, while a mid-task support
transfer between the two arms extends the effective workspace
without any external clamp. The pipeline achieves an 8/10 end-to-end
success rate, a cell-localisation root-mean-square error of
$2.4$\,mm, and a mean cycle time of 6.0\,minutes per pack, providing
a practical, fixture-free building block for industrial battery
recycling.
\end{abstract}


\end{frontmatter}

\section{Introduction}
\label{sec:intro}

The accelerating electrification of transport and consumer
electronics has established lithium-ion batteries as one of the most
strategically important industrial commodities of the past decade.
Worldwide sales of passenger electric vehicles (EVs) exceeded ten
million units in 2022 and are projected to roughly double within
the next several years~\cite{iea2023outlook}.
Given a typical service life of 8--15~years, the annual volume of
retired EV battery packs requiring safe and cost-effective
processing is expected to reach the order of several million units
by 2030, in addition to a continuous inflow of retired packs from
power tools, e-bikes, and portable
electronics~\cite{meng2022intelligent}.
End-of-life battery packs are not a uniform waste stream: they
contain valuable cathode metals (lithium, cobalt, nickel, and
manganese) whose primary supply is geographically concentrated, and
they pose substantial safety risks arising from residual charge,
flammable electrolytes, and the potential for thermal
runaway~\cite{harper2019recycling}.
Effective recycling therefore requires accurate, repeatable, and
safe disassembly to at least the module level, and ideally to the
individual cell, so that subsequent sorting and material recovery
can be carried out at high purity.

The prevailing solution is still manual disassembly.
Human operators accommodate the wide diversity of pack geometries
that has emerged across vehicle manufacturers, but they are also
exposed to toxic and flammable gases, high-voltage terminals, and
residual electrical charge throughout the
process~\cite{harper2019recycling,kaarlela2024review}.
Manual throughput is poorly matched to the projected scale of
end-of-life inflows and is fundamentally constrained by the safety
procedures that the task imposes.
These factors have motivated a growing body of research on robotic
battery disassembly as a complement to, or a partial replacement of,
manual production lines~\cite{villagrossi2023robotics,
meng2022intelligent,zang2024robotic}.

Robotic disassembly has been demonstrated at the pack, module, and
individual-cell levels of the product hierarchy, and a substantial
literature has accumulated on grasp pose estimation, visual servoing,
and dual-arm coordination directly relevant to the
problem~\cite{kuka2024demobat,kay2022robotic,
qu2024robotic,liang2025experimental,fang2020graspnet,tian2025fabrica}.
Across these prior systems, the majority of demonstrations rely on
one or more of three simplifying assumptions.
First, the pack is presented in a known, fixtured pose, removing the
need for vision-based localisation.
Second, the disassembly procedure terminates at the module level
rather than at individual cylindrical cells, circumventing the
sub-centimetre localisation accuracy that cell-level extraction
demands.
Third, cell extraction relies on specialised tooling, such as vacuum
suction or custom jigs, which restricts deployment flexibility.

This paper addresses the gap identified above.
The objective is to disassemble, to the level of individual
cylindrical cells, a battery pack of unknown initial position and
orientation, using a dual-arm robotic system equipped only with
general-purpose parallel-jaw grippers, commodity RGB-D sensors, and
a pre-trained grasp detector, and without recourse to any external
fixture or custom tooling.

The proposed solution is a vision-guided, multi-stage pipeline that
decomposes the disassembly task into four sequential stages: pack
acquisition, lid removal and cell dump, cell-assembly relocation, and
sequential cell extraction. Each stage is implemented by a
stage-specific policy that maps RGB-D observations to robot motion
commands, drawing on a pre-trained
GraspNet-Baseline~\cite{fang2020graspnet} detector for grasp
proposals, on look-and-move wrist-camera corrections in place of
continuous visual servoing,
and on a mid-task support transfer between the two arms that extends
the effective workspace without any external clamp.

The pipeline is implemented on a Realman~RM75 dual-arm platform with
DH~Robotics~AG95 grippers and three Intel RealSense~D435 RGB-D
cameras (one head-mounted, two wrist-mounted), running ROS~Noetic
and MoveIt~1. It is validated on a non-functional mock-up
dimensionally identical to a commercial 18650 module (21~cells in a
$3\times 7$ layout).
The main contributions of this paper are as follows:
\begin{enumerate}[label=(\roman*),leftmargin=*,itemsep=2pt]
  \item Cell-level 18650 extraction from a pack at an unknown
        initial pose is formulated as a perception-driven dual-arm
        manipulation task, and an end-to-end implementation is
        reported that relies only on standard parallel-jaw grippers
        and commodity RGB-D sensing, without dedicated fixtures or
        specialised tooling.
  \item A discrete-correction look-and-move perception scheme is
        developed that couples learned grasp proposals with a
        single-shot wrist-camera refinement step, offering a
        practical compromise between the accuracy of continuous
        visual servoing and the latency of open-loop planning, and
        extending the perception scope to the enclosure-opening and
        cell-localisation stages.
  \item A fixture-free dual-arm support-transfer strategy is
        introduced, in which the two arms exchange stabilising and
        extracting roles to substitute kinematic coordination for
        dedicated mechanical fixtures. The complete pipeline is
        evaluated on a physical platform, and the role of each
        perception component is examined through ablation studies.
\end{enumerate}

The remainder of the paper is organised as follows.
Section~\ref{sec:related} reviews related work and positions the
contribution.
Section~\ref{sec:system} presents the system architecture, problem
formulation, motion-planning framework, and hand-eye calibration.
Section~\ref{sec:pipeline} details the four-stage
perception-and-manipulation pipeline.
Section~\ref{sec:experiments} reports the experimental platform,
per-stage and end-to-end results, ablations, and a comparison with
related work.
Section~\ref{sec:discussion} discusses key findings, limitations, and
future directions, and Section~\ref{sec:conclusion} concludes.

\section{Related Work}
\label{sec:related}

\subsection{Robotic disassembly for battery recycling}
\label{sec:rel_disassembly}

The growing scale of end-of-life battery flows has driven a sharp
increase in research on robotised disassembly over the past five
years.
A recent systematic review by Kaarlela~et al.~\cite{kaarlela2024review}
covers 63 publications and concludes that the diversity of pack
designs and the safety risks of handling energised cells make fully
automated disassembly difficult.
The review identifies human-robot collaboration as the most practical
route to combine robotic repeatability with human flexibility.
Complementary reviews by
Villagrossi and Dinon~\cite{villagrossi2023robotics},
Meng~et al.~\cite{meng2022intelligent}, and
Zang~et al.~\cite{zang2024robotic} reach a similar conclusion
and consistently identify perception-guided operation and flexible
manipulation under pose uncertainty as the two central enabling
capabilities for further automation.

Existing disassembly demonstrations span all levels of the product
hierarchy.
At the pack level, the Fraunhofer DeMoBat~\cite{kuka2024demobat}
project uses industrial robots to remove bolts and cables from full
automotive packs whose geometry is known a priori, demonstrating
that high-payload automation is feasible when geometric uncertainty
is removed.
At the module level, Kay~et al.~\cite{kay2022robotic} combine
force-controlled insertion with safety interlocks for
cylindrical-cell extraction in a human-robot cell.
Qu~et al.~\cite{qu2024robotic} use four arms (two KUKA iiwa~14 and
two Techman~14) with two-step visual localisation to disassemble
plug-in-hybrid battery modules, while
Liang~et al.~\cite{liang2025experimental} report a techno-economic
analysis of a robotic prismatic-module cell whose productivity is
comparable to 5.5 human workers, and
Erdogan~et al.~\cite{contreras2024multirobot} verify RRT-based
collision-free planning for a simulated four-robot disassembly cell.
At the cell level, prior work is sparse.
To the best of our knowledge, cell-level extraction of 18650 cylindrical cells without fixtures has not been demonstrated in the literature

These works show that perception-guided robotic operation is
feasible, but they also illustrate the three simplifying assumptions
noted in the Introduction.
Cell-level extraction of cylindrical 18650 cells without external
fixtures remains under-explored: the tight packing of an 18650 array
(1\,mm edge-to-edge gap, 18\,mm cell diameter) demands sub-centimetre
localisation accuracy, which exceeds what fixed-camera setups
typically deliver.

\subsection{Perception for grasping and disassembly}
\label{sec:rel_perception}

Estimating feasible 6-DoF (six degrees of freedom) grasp poses from
raw sensor data is a prerequisite for manipulating objects whose
position and orientation are not known a priori, and the past decade
has seen rapid progress on data-driven solutions to this problem.
Mousavian~et al.~\cite{mousavian20196dof} were among the first
to formalise grasp prediction as a sampling-and-scoring problem,
using a variational autoencoder to generate candidate 6-DoF grasps
and a separate evaluator to score them.
GraspNet-1Billion~\cite{fang2020graspnet} scales this paradigm to a
large-scale benchmark with a dense point-cloud network that predicts
grasp candidates directly on visible surfaces.
Contact-GraspNet~\cite{sundermeyer2021contact} improves efficiency in
cluttered scenes by predicting contact surfaces, and
AnyGrasp~\cite{fang2023anygrasp} extends the approach to robust grasp
prediction across a wide range of unseen objects and viewpoints.
A consistent practical limitation of these networks is the domain
gap between the training distribution (diverse household objects,
predominantly textured and matte) and industrial targets such as
textureless, highly reflective battery enclosures.
Ma~et al.~\cite{ma2024generalizing} address this through
physical constraint regularisation and contact-score joint
optimisation, while fine-tuning for each new pack geometry remains
impractical at recycling scale due to annotation cost and the
non-stationary distribution of retired EV models.

A complementary line of work refines coarse perception with
visual feedback during execution.
Classical visual servoing distinguishes image-based (IBVS) and position-based (PBVS)
formulations. The latter reconstructs 3-D targets from depth data
and is better suited to metric manipulation tasks.
Both classically run as tight feedback loops at sensor frame rate,
which is incompatible with the per-call planning latency of
sampling-based motion planners.
A pragmatic alternative is the look-and-move variant, in which a
single correction is computed from a fresh wrist-camera observation
and the motion is executed open-loop to the corrected target.
Zhou~et al.~\cite{zhou2024towards} apply a coarse-to-fine vision
pipeline to screw disassembly, demonstrating the practical benefit
of an additional close-range correction step between coarse
detection and tool engagement.

For sub-tasks with a strong geometric prior, classical parametric
detectors remain attractive.
The circular end-caps of 18650 cells, for example, have a known
diameter, which makes a parametric detector such as HoughCircles a
natural match: it requires no training data, its parameters can be
derived analytically from camera intrinsics and scene depth, and
its noise characteristics admit straightforward multi-frame
averaging.
This suggests that, in industrial pipelines whose components have
known geometry, a hybrid learn-and-filter perception stack can
trade favourably against a uniformly learned alternative in terms
of deployment cost and maintainability.

\subsection{Dual-arm manipulation and coordination}
\label{sec:rel_dualarm}

Dual-arm systems extend the kinematic reach and dexterity of a single
robot, but introduce significant planning complexity from the higher
dimensionality of the joint space and the need for mutual collision
avoidance.
Recent work has addressed both ends of this trade-off.
Fabrica~\cite{tian2025fabrica} combines hierarchical task planning
with learned manipulation policies to demonstrate multi-part assembly
across diverse object geometries.
Assembly and disassembly differ structurally: assembly proceeds
toward a known target state against which progress can be verified,
whereas disassembly starts from an uncertain initial state and must
react to observations at each stage.
Grannen~et al.~\cite{grannen2023stabilize} demonstrate a
bimanual stabilise-and-act coordination scheme in which one arm
holds an object while the other performs precise manipulation,
illustrating how role specialisation can simplify bimanual planning.

Earlier dual-arm disassembly systems typically assume known,
fixture-aligned parts.
Huang~et al.~\cite{huang2021collaborative} present a
human-robot collaborative cell for press-fitted components but rely
on a human to position the parts precisely.

Fixture-free coordination strategies that extend the effective
workspace through inter-arm role exchange remain comparatively
under-explored in the disassembly setting.

Taken together, the three lines of work above leave a consistent gap.
Cell-level cylindrical disassembly without fixtures is rare, learned
grasp detectors are not by themselves accurate enough for the tight
clearances of 18650 arrays, and dual-arm coordination has mostly been
studied in assembly or fixtured settings rather than as a substitute
for external clamping. The pipeline introduced in the remainder of
this paper targets precisely this combination.

\section{System Overview and Preliminaries}
\label{sec:system}

This section introduces the system architecture and overall data
flow on which the pipeline is built (Section~\ref{sec:hw}), formalises
the disassembly task as a staged decision problem
(Section~\ref{sec:problem}), and describes the motion-planning
(Section~\ref{sec:motion}), and eye-in-hand calibration with
depth-to-base-frame projection (Section~\ref{sec:calib}) components
that the four-stage perception-and-manipulation pipeline of
Section~\ref{sec:pipeline} relies on.

\subsection{System architecture}
\label{sec:hw}

\begin{figure}[!t]
\centering
\includegraphics[width=0.95\linewidth]{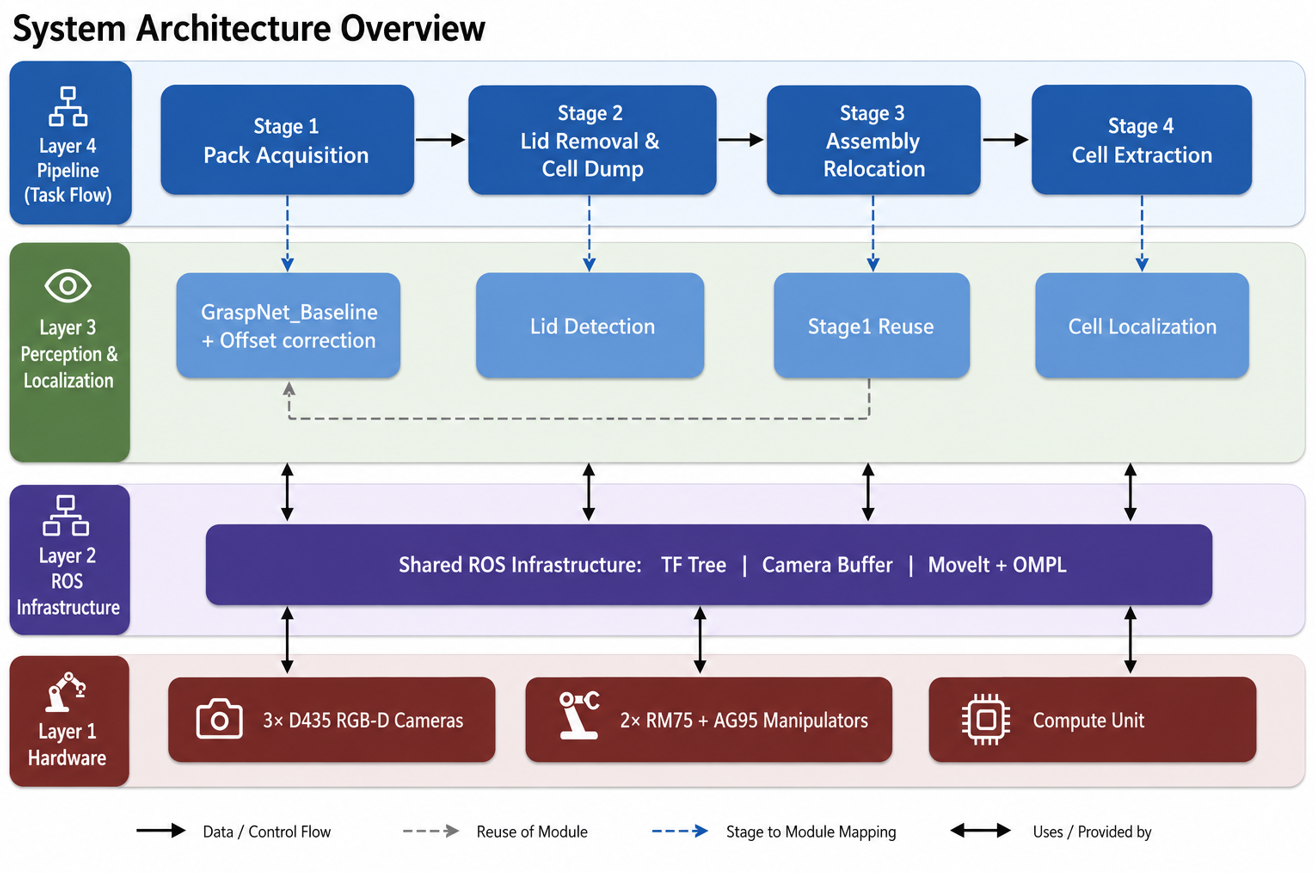}
\caption{Overall system architecture. Three RGB-D streams feed a
learned 6-DoF grasp detector and classical depth/colour perception
modules. The resulting targets are consumed by a shared
motion-planning layer that drives the two arms and their grippers
through a common TF tree.}
\label{fig:system_overview}
\end{figure}

The proposed system follows a sensing, perception, planning, and
actuation pipeline distributed across two cooperating seven
degrees-of-freedom (7-DoF) arms, each fitted with a parallel-jaw
electric gripper, and three RGB-D cameras with complementary
viewpoints (Figure~\ref{fig:system_overview}).
A head-mounted camera provides a wide, fixed-baseline view of the
workspace and supplies the point clouds consumed by a learned 6-DoF
grasp detector, which produces coarse grasp candidates for the
arbitrarily posed pack.
Each arm additionally carries a wrist-mounted camera that delivers
close-range, viewpoint-independent observations of the active target.
These feed classical perception modules (a depth-and-colour fused
lid detector and a Hough-circle cell localiser) whose outputs serve
as discrete look-and-move corrections immediately before each
precision action.
All perception outputs converge on a shared motion-planning layer
that maintains a single transform (TF) tree, performs collision-aware
trajectory generation, and issues joint and gripper commands to the
two arms.

This three-camera configuration reflects a deliberate division of
perceptual labour.
The head camera supplies the wide context required to bootstrap
learned grasp detection from an unknown initial pack pose, but its
viewing geometry and depth noise at table distance are inadequate
for the sub-centimetre alignment required for cell extraction.
The wrist cameras absorb the residual error by approaching each
target before action.
On the actuation side, the dual-arm layout enables a fixture-free
support-transfer strategy in which the two arms exchange stabilising
and extracting roles to extend the effective workspace, supported by
three motion-planning modes (single-arm, parallel, and simultaneous
dual-arm planning) that are invoked selectively at each stage
(Section~\ref{sec:motion}).

\subsection{Problem formulation}
\label{sec:problem}

We consider a tabletop disassembly task in which an intact, arbitrarily
posed 18650 pack must be processed into its constituent cylindrical cells
and deposited into output bins.
Let $\mathcal{W}\subset\mathbb{R}^{3}$ be the visible workspace, and
let $\mathcal{B}_0$ denote the initial pack configuration (position,
orientation, internal assembly state).
The combined joint configuration of the two 7-DoF arms is
$\bar{\mathbf{q}}=[\mathbf{q}^{L\top},\mathbf{q}^{R\top}]^{\top}\in\mathbb{R}^{14}$.
An observation at time $t$ aggregates the active cameras
$\mathcal{A}_t$ as
\begin{equation}
\mathcal{O}_t = \bigl\{\bigl(\mathbf{I}^{\mathrm{rgb}}_{c,t},\,
                            \mathbf{I}^{\mathrm{d}}_{c,t},\,
                            \mathbf{K}_c\bigr)\;\big|\; c\in\mathcal{A}_t\bigr\},
\label{eq:observation}
\end{equation}
where $\mathbf{I}^{\mathrm{rgb}}$, $\mathbf{I}^{\mathrm{d}}$, and
$\mathbf{K}$ are the RGB image, aligned depth image, and intrinsic matrix
of camera~$c$.
An action is a pair of end-effector targets together with gripper
commands,
\begin{equation}
a_t = \bigl(\mathbf{T}^L_t,\,\mathbf{T}^R_t,\, g^L_t,\, g^R_t\bigr),
\quad \mathbf{T}^{\{L,R\}}_t\in SE(3),\;
g^{\{L,R\}}_t \in [0,1].
\label{eq:action}
\end{equation}
Single-arm steps set the idle arm's target to its current pose.

The task is decomposed into four sequential stages, and the
corresponding stage policies $\pi_k$ map observations to actions,
\begin{equation}
a_t = \pi_k(\mathcal{O}_t;\theta_k),\quad k=\text{stage}(t),
\label{eq:staged}
\end{equation}
where $\theta_k$ collects the (frozen) GraspNet weights and the
hand-tuned thresholds.
GraspNet is the only learned component, and all remaining perception
and control logic are rule-based.
Table~\ref{tab:stage_overview} summarises the stages.

\begin{table}[!t]
\centering
\caption{Overview of the four-stage disassembly pipeline.}
\label{tab:stage_overview}
\small
\begin{tabular}{clll}
\toprule
\textbf{Stage} & \textbf{Name} & \textbf{Arm(s)} & \textbf{Completion condition} \\
\midrule
1 & Battery pack acquisition   & Left (+ Right parallel) & Pack held at fixed position \\
2 & Lid removal \& cell dump   & Right + Left            & Cells deposited on workspace \\
3 & Cell assembly relocation   & Left                    & Holder seated at disassembly position \\
4 & Sequential cell extraction & Both                    & All cells in output bins \\
\bottomrule
\end{tabular}
\end{table}

Between Stages~2 and~3, the operator removes the thin upper holder
bracket ($5$--$10$\,s). This part is friction-fit and cannot
be reliably grasped by a parallel-jaw gripper.
This is the only manual step in the pipeline.

\subsection{Motion planning and control}
\label{sec:motion}

MoveIt~1 on ROS~Noetic exposes the dual-arm robot through three move
groups: \texttt{l\_arm} and \texttt{r\_arm} for independent single-arm
planning, and \texttt{dual\_arms} for simultaneous 14-DoF planning.
The primary planner is
RRT-Connect~\cite{kuffner2000rrtconnect}, chosen for its bidirectional
growth and proven performance on 7-DoF arms operating close to the
tabletop.
Path optimality is secondary to planning speed in this setting, because
trajectories are generated at runtime from the current state and the
workspace is relatively open.
On top of this baseline, three modifications are introduced to
improve reliability for the present task.

The first modification addresses inverse-kinematics selection.
For a kinematically redundant 7-DoF arm, the KDL inverse-kinematics
solver~\cite{smits2011kdl} can return very different joint
configurations across calls for the same target pose, occasionally
producing unnecessarily large joint displacements.
To stabilise this behaviour, multiple solutions are requested (up to
$N=8$ attempts) and the one closest in joint space to the current
configuration is selected, with an early-stop threshold
$d_\text{stop}=0.5$\,rad (Algorithm~\ref{alg:nearest_ik}).
The selected $\mathbf{q}^*$ is then used as the seed for a second
joint-space planning query, decoupling the kinematic solve from
trajectory planning so that each can use its own parameter set.

\begin{algorithm}[H]
\DontPrintSemicolon
\caption{Nearest-IK multi-solution selection.}
\label{alg:nearest_ik}
\KwIn{Move group \texttt{arm}; target pose $\mathbf{T}_\text{des}$;
      attempts $N{=}8$; early-stop $d_\text{stop}{=}0.5$\,rad}
\KwOut{Nearest joint configuration $\mathbf{q}^*$ and distance $d^*$}
$\mathbf{q}_\text{cur}\leftarrow$ current joints;\;
$\mathbf{q}^*\leftarrow\textsc{None}$;\quad $d^*\leftarrow\infty$\;
\For{$i=1,\ldots,N$}{
  $(\text{ok},\tau_i)\leftarrow\texttt{arm.plan}(\mathbf{T}_\text{des})$
    \tcp*{random IK seed}
  \If{$\text{ok}$ \textbf{and} $\tau_i$ has waypoints}{
    $\mathbf{q}_i\leftarrow$ final waypoint of $\tau_i$;\;
    $d_i\leftarrow\|\mathbf{q}_i-\mathbf{q}_\text{cur}\|_2$;\;
    \If{$d_i<d^*$}{
      $d^*\leftarrow d_i$;\quad $\mathbf{q}^*\leftarrow\mathbf{q}_i$;\;
      \lIf{$d^*<d_\text{stop}$}{\textbf{break}}
    }
  }
}
\Return $(\mathbf{q}^*,d^*)$\;
\end{algorithm}

The second modification is a joint-space fallback for Cartesian
planning.
Approach, lift, and press-down operations use MoveIt's Cartesian
interpolation, with a small step size ($\delta=5$\,mm) near the
object and a larger one ($\delta=10$\,mm) for transport.
The planner returns the fraction $f\in[0,1]$ of the requested path
that is feasible.
When $f<f_\text{accept}=0.8$, the Cartesian path is abandoned and the
system falls back to a joint-space plan to the same target.
Straight-line accuracy is sacrificed, but the trajectory remains
collision-free and the task can complete.
This fallback is the dominant source of robustness against the
Cartesian-planning anomalies that would otherwise abort a stage.

The third modification concerns dual-arm coordination, for which the
pipeline uses three modes that trade off planning cost against safety.
Separate planning moves a single arm at a time using either
\texttt{l\_arm} or \texttt{r\_arm}, and is the default.
Parallel execution plans the two arms independently and executes the
resulting trajectories simultaneously.
This overlaps single-arm motions but does not guarantee mutual
collision avoidance, and is therefore restricted to phases in which
the two motions are known to be spatially disjoint.
Simultaneous planning operates in the 14-DoF \texttt{dual\_arms} group
and guarantees mutual avoidance at the cost of longer planning times
and occasional planning failures. It is therefore reserved for steps
in which the two arms must coordinate closely, such as the final
dual-arm release in Stage~4.

\subsection{Hand-eye calibration and depth projection}
\label{sec:calib}

Each wrist camera requires a constant transform
$\mathbf{X}=\mathbf{T}_{\mathcal{C}}^{\mathcal{E}}\in SE(3)$, obtained by
solving the standard eye-in-hand equation
$\mathbf{A}_i\mathbf{X}=\mathbf{X}\mathbf{B}_i$.
An ArUco marker (DICT\_5X5\_50, ID~2, 50\,mm) is observed from
$n\geq 10$ diverse arm poses. The marker pose is recovered with
\texttt{cv2.solvePnP}, and the end-effector pose comes from the MoveIt
TF tree.
OpenCV's five closed-form solvers are run in parallel. Candidates with $|\det(\mathbf{R})-1|\geq 0.1$ or
$\|\mathbf{t}\|\geq 0.5$\,m are rejected, and the result with the
largest median translation norm is selected, because the median is
more robust than the mean to the degenerate failure modes observed
empirically.
The chosen $\mathbf{X}$ is accepted when the per-axis standard
deviation of the back-projected marker is below 5\,mm across all
$n$~samples, otherwise additional samples are collected.

Table~\ref{tab:calib_results} reports the calibration results.
All three cameras pass the 5\,mm threshold.
Wrist-camera translation norms (60--70\,mm) are consistent with the
physical mounting offset, and the head-camera value of approximately
413\,mm reflects the overhead mounting distance.

\begin{table}[!t]
\centering
\caption{Hand-eye calibration results for the three D435 cameras.
$n$: number of calibration samples; method: winning closed-form solver
among \{TSAI, PARK, HORAUD, ANDREFF, DANIILIDIS\};
$\|\mathbf{t}\|$: translation norm; $\sigma_{xy},\sigma_z$: per-axis
back-projection std.\ dev.\ on held-out poses.}
\label{tab:calib_results}
\small
\begin{tabular}{lccccc}
\toprule
\textbf{Camera} & $n$ & \textbf{Method} &
  $\|\mathbf{t}\|$ (mm) & $\sigma_{xy}$ (mm) & $\sigma_z$ (mm) \\
\midrule
Head (eye-to-hand) & 18 & TSAI    & 412.8 & 3.5 & 2.8 \\
Left wrist         & 15 & PARK    & 63.2  & 2.1 & 1.8 \\
Right wrist        & 12 & HORAUD  & 67.5  & 2.4 & 2.0 \\
\bottomrule
\end{tabular}
\end{table}

With $\mathbf{X}$ in hand, a point $(u,v)$ in a depth map projects
into the camera frame through the pinhole model and into the base
frame through the kinematic chain:
\begin{equation}
\mathbf{p}_{\mathcal{C}}=\Bigl[\tfrac{(u-c_x)d}{f_x},\;
                                \tfrac{(v-c_y)d}{f_y},\; d\Bigr]^{\top},
\quad
\tilde{\mathbf{p}}_{\mathcal{B}}=
  \mathbf{T}_{\mathcal{E}}^{\mathcal{B}}(\mathbf{q}_t)\,
  \mathbf{T}_{\mathcal{C}}^{\mathcal{E}}\,
  \tilde{\mathbf{p}}_{\mathcal{C}},
\label{eq:projection}
\end{equation}
where $d=\mathbf{I}^{\mathrm{d}}(u,v)$ and $\tilde{\mathbf{p}}$ is the
homogeneous form.
For the fixed head camera, $\mathbf{T}_{\mathcal{C}}^{\mathcal{E}}$ is
replaced by the static extrinsic $\mathbf{T}_{\mathcal{C}}^{\mathcal{B}}$.
Equation~\eqref{eq:projection} is invoked by every perception module
in Section~\ref{sec:pipeline} with the appropriate pixel set and
camera.

\section{Disassembly Pipeline}
\label{sec:pipeline}

\subsection{Pipeline overview}
\label{sec:pipeline_overview}

\begin{figure}[!t]
\centering
\begin{subfigure}[t]{0.48\linewidth}
  \centering
  \includegraphics[width=\linewidth,height=0.75\linewidth,keepaspectratio]{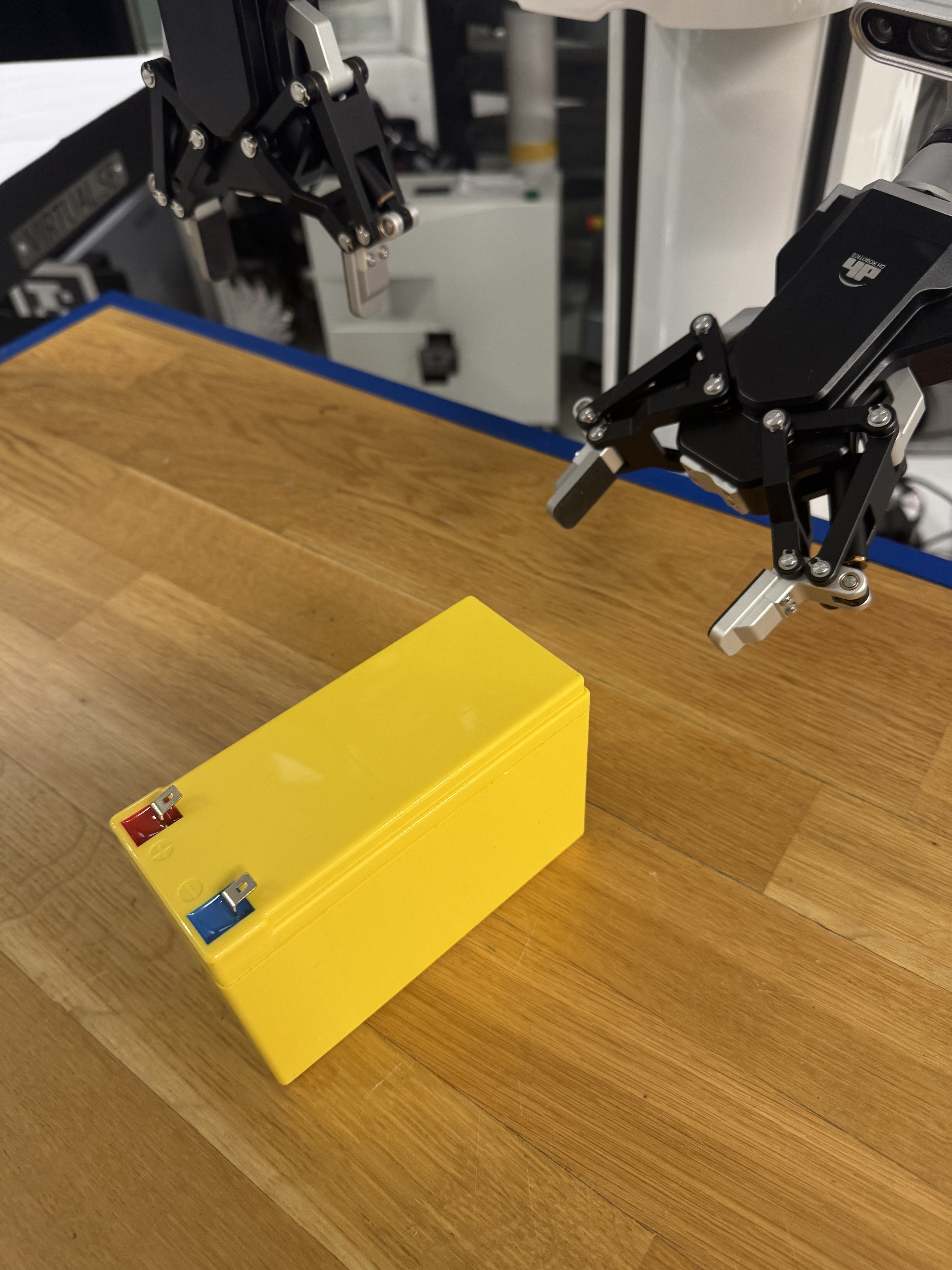}
  \caption{Stage~1: left arm at the pre-grasp pose.}
\end{subfigure}\hfill
\begin{subfigure}[t]{0.48\linewidth}
  \centering
  \includegraphics[width=\linewidth,height=0.75\linewidth,keepaspectratio]{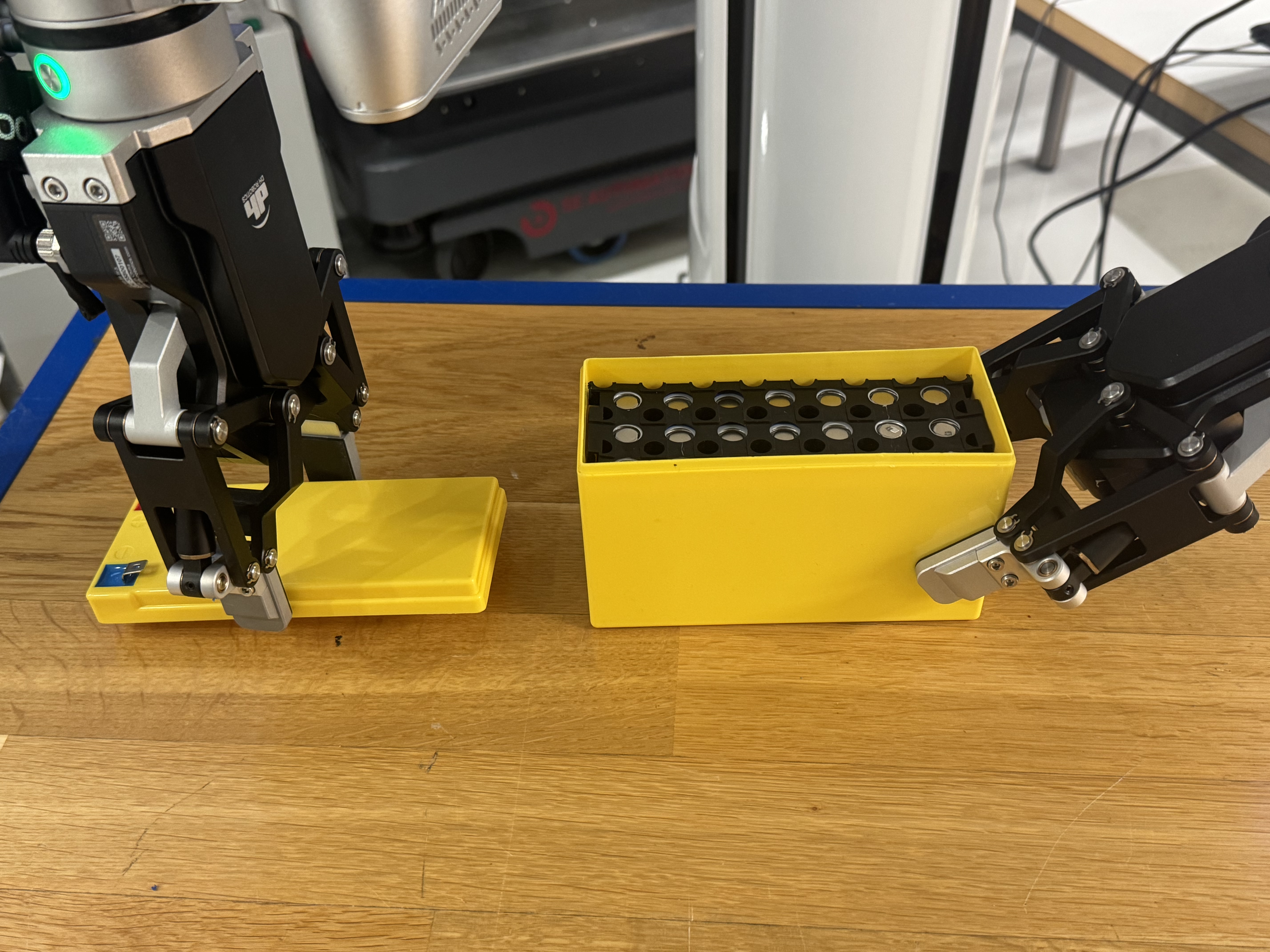}
  \caption{Stage~2: lid removed, 21 cells exposed.}
\end{subfigure}\\[0.2cm]
\begin{subfigure}[t]{0.48\linewidth}
  \centering
  \includegraphics[width=\linewidth,height=0.75\linewidth,keepaspectratio]{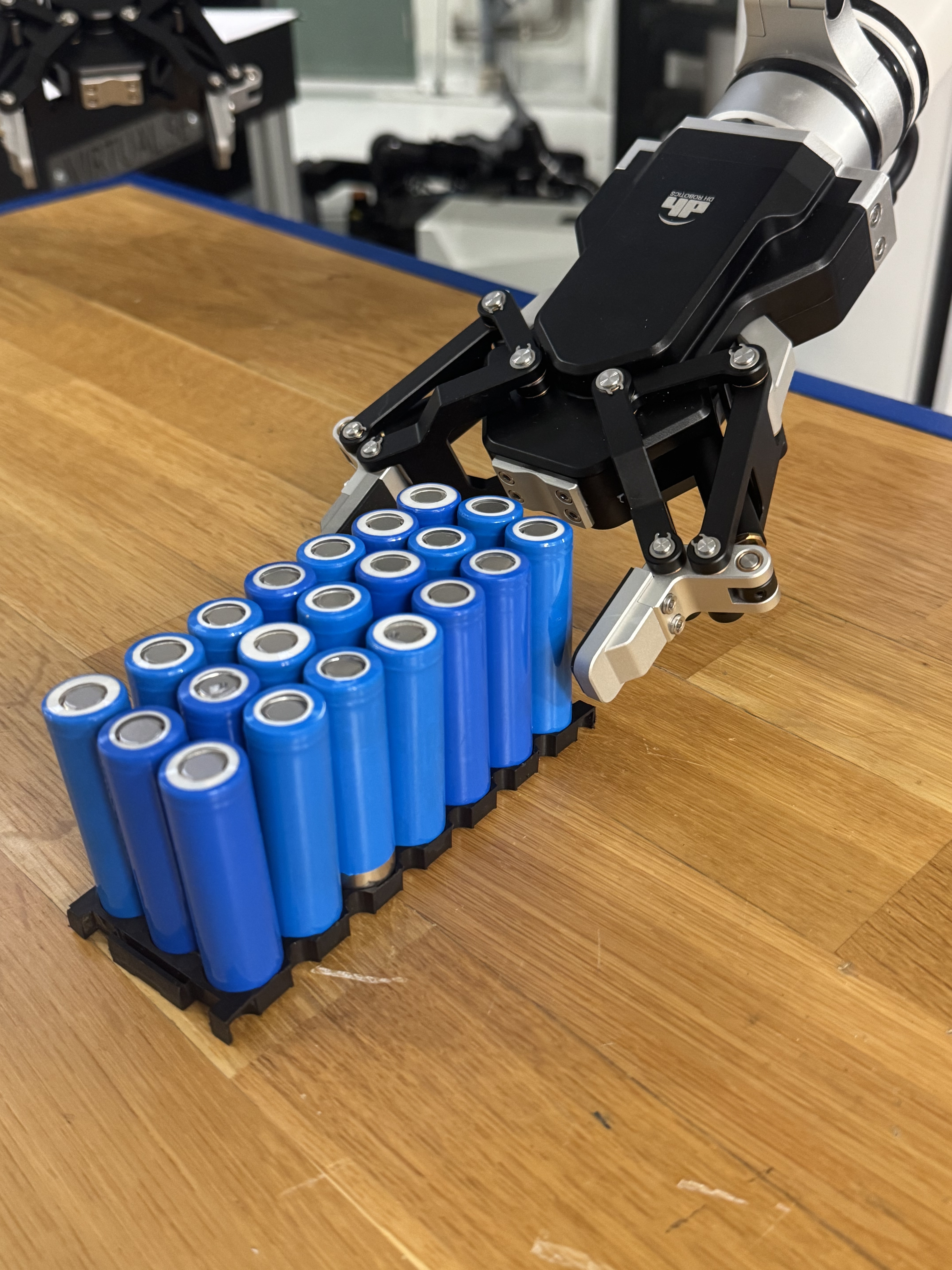}
  \caption{Stage~3: assembly grasp prior to relocation.}
\end{subfigure}\hfill
\begin{subfigure}[t]{0.48\linewidth}
  \centering
  \includegraphics[width=\linewidth,height=0.75\linewidth,keepaspectratio]{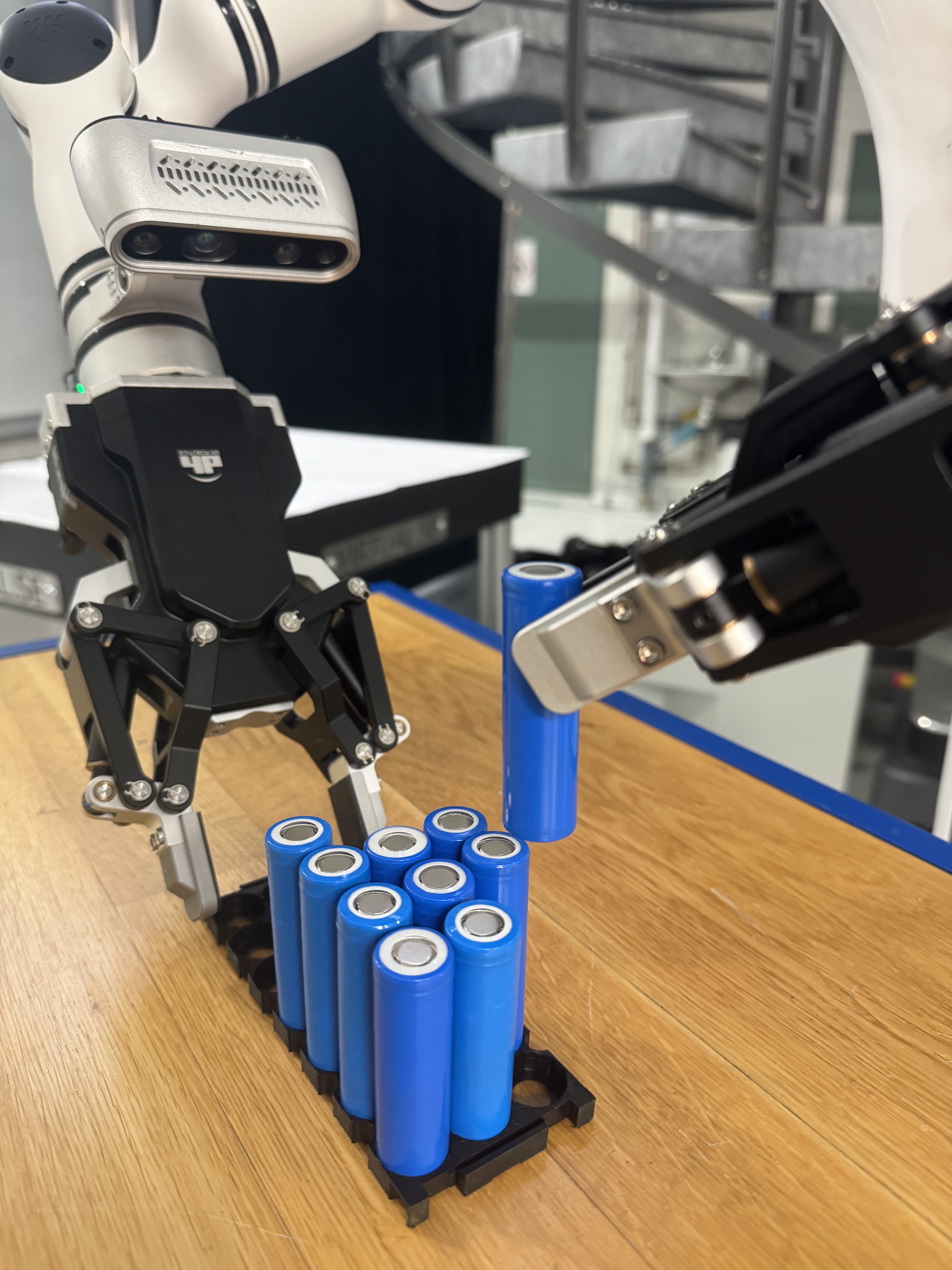}
  \caption{Stage~4: sequential cell extraction.}
\end{subfigure}
\caption{Snapshots of the four pipeline stages.}
\label{fig:pipeline_overview}
\end{figure}

This section describes the four stages of the pipeline in detail.
Algorithm~\ref{alg:pipeline} gives the top-level control flow that
the rest of the section unfolds;
Figure~\ref{fig:pipeline_overview} shows representative platform
snapshots from each stage.
For readability, all symbolic parameters
($N_\text{warm}$, $\epsilon_k$, $h_\text{lift}$, $N_\text{avg}$,
$\delta_z$, $D_\text{max}$, etc.) are introduced in context here and
their numerical values, together with the rationale behind each
value, are consolidated in Section~\ref{sec:experiments}.

\begin{algorithm}[H]
\DontPrintSemicolon
\caption{Top-level disassembly pipeline.}
\label{alg:pipeline}
\KwIn{Pack $\mathcal{B}_0$ at unknown pose in workspace $\mathcal{W}$}
\KwOut{All $M$ cells in output bins $\mathcal{D}$}
Initialise both arms to home; open grippers; warm up cameras
  (discard $N_\text{warm}$ frames)\;
\tcp{Stage~1: Pack acquisition}
$g^*\leftarrow\textsc{GraspSelect}(\mathcal{O}_\text{head},\mathcal{W}_1,\epsilon_1)$
  \tcp*{Algorithm~\ref{alg:graspselect}}
Apply lateral correction $\Delta x_{\mathcal{B}}$ from left wrist camera\;
Approach $\to$ grasp $\to$ pitch-up $\to$ lift $h_\text{lift}$
  $\to$ transport to $\mathbf{p}_\text{fix}$
  $\to$ open $\to$ press-down $\to$ re-grip (full force)\;
[parallel] right arm $\to$ overhead observation pose\;
\tcp{Stage~2: Lid removal and cell dump}
$[z_\text{lid},\mathbf{c}_\text{lid},\psi_\text{lid}]
   \leftarrow\textsc{LidDetect}(\mathcal{O}_\text{wrist-R},N_\text{avg})$
  \tcp*{depth+HSV+PCA}
Right: $Y$-align $\to$ rotate $\psi_\text{lid}$ $\to$ descend to
$z_\text{lid}$ $\to$ grasp $\to$ lift $\to$ bin\;
Left: pitch-up $\to$ flip $+180^{\circ}$ $\to$ tilt $\to$
release cell assembly\;
\tcp{Manual step: operator removes upper holder bracket}
\tcp{Stage~3: Assembly relocation}
$g^*\leftarrow\textsc{GraspSelect}(\mathcal{O}_\text{head},\mathcal{W}_3,\epsilon_3)$\;
Grasp (reduced force) $\to$ lift $\to$ transport to $\mathbf{p}_\text{fix}$
$\to$ position-controlled press-down ($\delta_z$, max $D_\text{max}$)\;
\tcp{Stage~4: Sequential cell extraction}
$\textsc{ExtractCells}(\text{right arm})$ \tcp*{Phase B}
Support transfer: right clamps holder; left releases\;
$\textsc{ExtractCells}(\text{left arm})$ \tcp*{Phase C}
Dual-arm simultaneous final release\;
\end{algorithm}

\subsection{Stage 1: Pack acquisition}
\label{sec:stage1}

Stage~1 begins with GraspNet inference and feasibility filtering.
Both arms move to home, the grippers open, and the cameras warm up
by discarding their first $N_\text{warm}$ frames so that
auto-exposure and white balance settle.
A point cloud is then constructed from the head-camera RGB-D stream,
masked by the Stage~1 workspace bounding box $\mathcal{W}_1$, and
randomly subsampled to $N_\text{pts}=20\,000$ points before being
passed to
GraspNet-Baseline~\cite{fang2020graspnet}, which is built on the
PointNet~\cite{qi2017pointnet} architecture.
The network returns a ranked set of 6-DoF proposals
$\mathcal{G}=\{g_i\}_{i=1}^{N}$ with confidence scores
$s_i\in[0,1]$, from which the highest-scoring feasible candidate
is selected as
\begin{equation}
g^* = \underset{g\in\mathcal{G}_\text{valid}}{\arg\max}\;\operatorname{score}(g),
\label{eq:graspopt}
\end{equation}
subject to workspace, approach-angle, and quality constraints,
\begin{equation}
\mathcal{G}_\text{valid}=\bigl\{g\in\mathcal{G}\,\big|\;
  \mathbf{p}_g\in\mathcal{W}_k,\;
  |a_z(g)|\leq\epsilon_k,\;
  \operatorname{score}(g)\geq\tau\bigr\},
\label{eq:graspfeasible}
\end{equation}
where $\mathbf{p}_g$ is the grasp centre in the base frame,
$a_z(g)$ the $z$-component of the approach direction, $\epsilon_k$
the approach-angle tolerance, and $\tau$ the minimum score.
A front-surface bias further restricts the candidate set to those
grasps within a distance $d_\text{front}$ of the maximum-$y$ point of
the cloud, targeting the accessible front face of the pack.

A PCA-based yaw refinement then sharpens the orientation.
The selected $g^*$ inherits its yaw from the network's prediction,
which on textureless enclosures is often noisy.
To improve it, principal component analysis is applied to the
$xy$-projection of the points within a radius $r_\text{pca}=100$\,mm
of the grasp centre, yielding eigenvectors $\mathbf{v}_1$ (major
axis) and $\mathbf{v}_2$ (minor axis) of the local front surface.
The gripper yaw is then set to align the closing direction with the
short-side direction of the enclosure,
\begin{equation}
\psi^* = \operatorname{atan2}(v_{2,y},v_{2,x}),
\label{eq:yaw_pca}
\end{equation}
with a sign disambiguation against the approach vector.
Algorithm~\ref{alg:graspselect} summarises the full selection
procedure.

\begin{algorithm}[H]
\DontPrintSemicolon
\caption{GraspNet-based grasp selection with PCA yaw alignment.}
\label{alg:graspselect}
\KwIn{RGB-D observation $\mathcal{O}$; workspace $\mathcal{W}_k$;
      approach threshold $\epsilon_k$; score threshold $\tau$}
\KwOut{Grasp $g^*$ with refined yaw $\psi^*$; or \textsc{Failure}}
$\mathcal{P}\leftarrow$ point cloud from $\mathcal{O}$ masked to $\mathcal{W}_k$;\;
$\mathcal{P}\leftarrow$ subsample to $N_\text{pts}$ points;\;
$\mathcal{G}\leftarrow\textsc{GraspNet}(\mathcal{P})$ \tcp*{ranked candidates}
$\mathcal{G}_\text{valid}\leftarrow\{g\in\mathcal{G}\mid$
  $\mathbf{p}_g\in\mathcal{W}_k,\;
  |a_z(g)|\leq\epsilon_k,\;
  \text{score}(g)\geq\tau\}$\;
$\mathcal{G}_\text{front}\leftarrow\{g\in\mathcal{G}_\text{valid}\mid
  (p_g)_y\geq\max_{\mathcal{P}}y-d_\text{front}\}$
  \tcp*{front-surface bias}
\If{$\mathcal{G}_\text{front}=\emptyset$}{retry up to 3 times;
  if still empty \Return \textsc{Failure}\;}
$g^*\leftarrow\arg\max_{g\in\mathcal{G}_\text{front}}\text{score}(g)$;\;
$\mathcal{P}_\text{loc}\leftarrow\{p\in\mathcal{P}\mid
  \|p_{xy}-p_{g^*,xy}\|\leq r_\text{pca}\}$;\;
$[\mathbf{v}_1,\mathbf{v}_2]\leftarrow\textsc{PCA}(\mathcal{P}_\text{loc} \text{ projected to } xy)$;\;
$\psi^*\leftarrow\operatorname{atan2}(v_{2,y},v_{2,x})$;\;
Disambiguate sign of $\psi^*$ against the approach vector
$\mathbf{a}_{g^*}$;\;
\Return $(g^*,\psi^*)$;\;
\end{algorithm}

\begin{figure}[!t]
\centering
\begin{subfigure}[t]{0.48\linewidth}
  \centering
  \includegraphics[width=\linewidth]{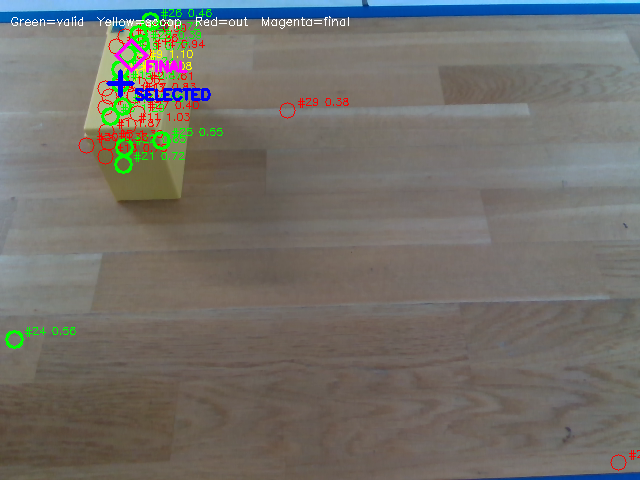}
  \caption{Workspace-filtered GraspNet candidates.}
\end{subfigure}\hfill
\begin{subfigure}[t]{0.48\linewidth}
  \centering
  \includegraphics[width=\linewidth]{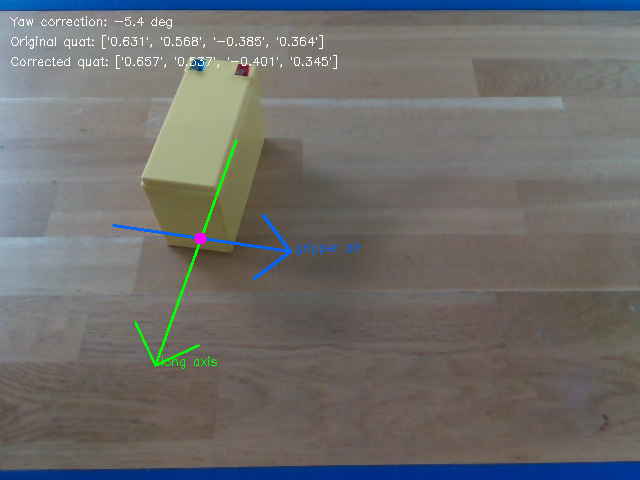}
  \caption{PCA eigenvectors: green $\mathbf{v}_1$ (major), blue
  $\mathbf{v}_2$ (minor, used for yaw).}
\end{subfigure}
\caption{Stage~1 perception. (a)~GraspNet candidates after
workspace/approach-angle filtering. (b)~PCA-based yaw alignment on the
pack front face.}
\label{fig:stage1_perception}
\end{figure}

A wrist-camera lateral correction then compensates for residual
head-camera error.
The grasp $g^*$ recovered from the head camera is precise enough in
depth and yaw but tends to drift laterally by several millimetres.
Three error sources contribute to this drift: range-dependent D435
depth noise propagated through the head-camera standoff of
approximately 400\,mm, residual reprojection error from the
head-camera hand-eye calibration itself ($\sigma_{xy}=3.5$\,mm,
Table~\ref{tab:calib_results}), and specular reflections on the
smooth plastic enclosure that degrade the front-face point-cloud
quality.
The combined offset can exceed the parallel-jaw capture range, and
the ablation in Section~\ref{sec:ablation} confirms the effect:
without correction, the Stage~1 success rate drops from 10/10 to
7/10.

To absorb this residual error, a single-shot look-and-move update
is performed once the left arm has reached the pre-grasp pose.
Its wrist camera captures one depth frame. The front surface of the
pack is then extracted by depth gating, and the lateral column
centroid $u_\text{cen}$ of the resulting mask is compared against a
pre-calibrated reference column
\begin{equation}
u_\text{ref}=\tfrac{W}{2}+\delta_u,
\label{eq:xcorr_ref}
\end{equation}
in which $\delta_u$ accounts for the lateral mounting offset of the
wrist camera relative to the tool centre point (TCP).
This offset is measured once by a one-time calibration experiment,
in which a marked object is grasped at its centre and the wrist
image is examined to read the pixel column on which the mark
appears.
The offset is empirically stable across the arm configurations used
in the experiments (108\,px on this platform).
The induced lateral correction in the base frame is
\begin{equation}
\Delta x_{\mathcal{B}}=-\frac{(u_\text{cen}-u_\text{ref})\,\bar{d}}{f_x},
\label{eq:xcorr_model}
\end{equation}
where $\bar{d}$ is the mean depth of the masked pixels and $f_x$ the
horizontal focal length.
A symmetric $\pm 10$\,mm deadband suppresses sensor noise. Whenever
the magnitude of the correction exceeds the deadband, the grasp
target is updated by $\Delta x_{\mathcal{B}}$ and the arm re-executes
the pre-grasp approach to the corrected position.

\begin{figure}[!t]
\centering
\includegraphics[width=0.65\linewidth]{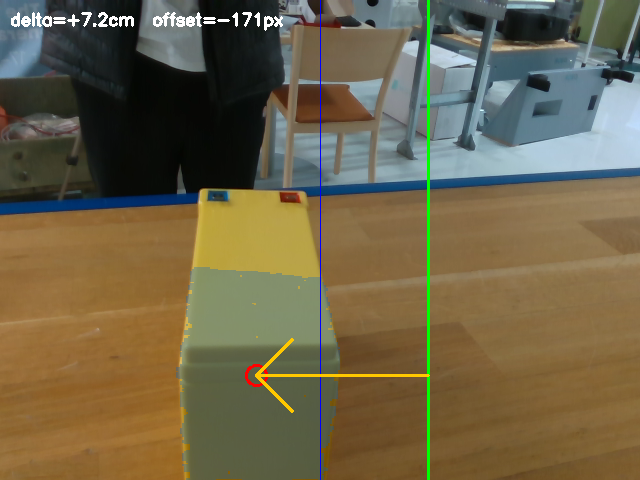}
\caption{Left-wrist lateral correction. Depth gating isolates the pack
front surface, and the orange arrow shows the measured offset
$\Delta x_{\mathcal{B}}$ between the column centroid $u_\text{cen}$ and
the calibrated reference $u_\text{ref}$.}
\label{fig:lateral_correction}
\end{figure}

Transport and stabilisation conclude the stage.
With the corrected target in hand, the left arm closes its gripper,
pitches up to clear the worktable, transports the pack to the fixed
disassembly position $\mathbf{p}_\text{fix}$, opens the gripper to
let the pack settle under gravity, descends a short distance, and
finally re-grasps at full force.
The descend-and-re-grasp step removes any residual error introduced
during transport, so that the pack starts Stage~2 from a repeatable
position and orientation.
In parallel, the right arm moves to the overhead observation pose
required by Stage~2. Because the two trajectories are spatially
disjoint, this overlap is handled by the parallel-execution mode
described in Section~\ref{sec:motion}.

\subsection{Stage 2: Lid removal and cell dump}
\label{sec:stage2}

Lid detection in Stage~2 uses a fused depth and colour pipeline.
With the pack secured at $\mathbf{p}_\text{fix}$, the right wrist
camera acquires $N_\text{avg}=30$ frames from the overhead pose and
runs a per-frame detection routine.
Each frame produces a binary mask by joint depth and HSV gating,
\begin{equation}
\mathbf{M}_f(u,v)=\mathbf{1}\!\bigl[
  d_\text{min}\!\leq\! d_f(u,v)\!\leq\! d_\text{max}\;\wedge\;
  \mathbf{h}_f(u,v)\in\mathcal{H}_\text{lid}\bigr],
\label{eq:lid_mask}
\end{equation}
with the depth range $[d_\text{min},d_\text{max}]$ chosen to match the
wrist-to-lid standoff at the topdown pose, and the HSV region
$\mathcal{H}_\text{lid}$ matched to the lid colour.
Frames with too few mask pixels are discarded.
For each valid frame, PCA on the mask pixels yields the image-plane
short axis. Two points along this axis are back-projected with
Eq.~\eqref{eq:projection} to obtain the corresponding base-frame
direction, from which the lid yaw is
\begin{equation}
\psi^{(f)}_\text{lid}=\operatorname{atan2}(e_{s,y}^{\mathcal{B}},e_{s,x}^{\mathcal{B}}).
\end{equation}
After rejecting frames whose height estimate deviates from the running
median by more than $\delta_z^\text{gate}=2$\,mm, the yaw is aggregated
by circular averaging,
\begin{equation}
\hat{\psi}_\text{lid}=
\operatorname{atan2}\!\Bigl(
  \tfrac{1}{|\mathcal{I}|}\!\sum_{f\in\mathcal{I}}\sin\psi^{(f)}_\text{lid},
  \tfrac{1}{|\mathcal{I}|}\!\sum_{f\in\mathcal{I}}\cos\psi^{(f)}_\text{lid}
\Bigr),
\label{eq:lid_circular_mean}
\end{equation}
which avoids $\pm\pi$ discontinuities. The lid centre and height are
the inlier means.
The output triple
$(\hat{z}_\text{lid},\hat{\mathbf{c}}_\text{lid},\hat{\psi}_\text{lid})$
fully parameterises the lid-removal grasp.

\begin{figure}[!t]
\centering
\includegraphics[width=0.65\linewidth]{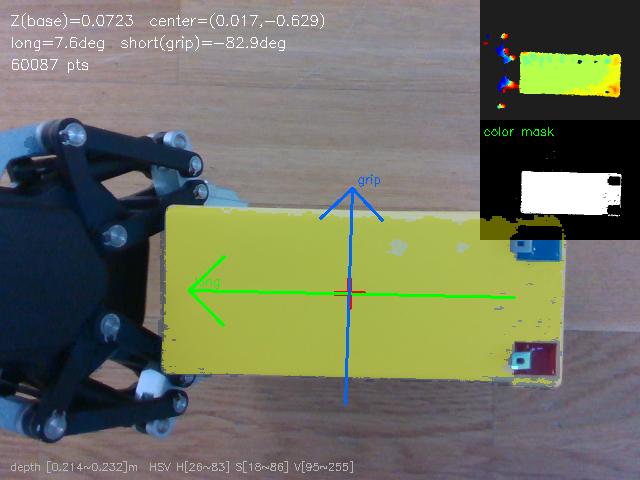}
\caption{Right-wrist lid detection: depth-gated and HSV-filtered mask
isolates the yellow lid, and the cross-hair marks the centroid while the
overlaid axes the PCA short axis used for yaw alignment.}
\label{fig:lid_detection}
\end{figure}

Lid removal and a gravity-assisted cell dump then follow.
Given the detection result, the right arm aligns to
$\hat{\mathbf{c}}_\text{lid}$, rotates its wrist by
$\hat{\psi}_\text{lid}-\psi_0$ to align the closing direction with
the lid's short axis, descends to $\hat{z}_\text{lid}$, grips at
maximum force, lifts vertically, and deposits the lid in a bin.
The left arm then pitches up, lifts, and flips the pack by
$+180^{\circ}$ to invert the cell assembly.
The flip-and-tilt motion is shaped by two physical considerations.
First, before the flip, the pitch is adjusted so that the contact
centroid moves from the fingertips towards the gripper base.
This increases the normal force and prevents the inverted pack from
slipping out of the gripper during the rotation.
Second, after the flip, a small additional pitch combined with a
10\,\% grip-force reduction allows the cell assembly to slide out
under gravity while residual friction continues to hold the outer
shell.
This exploits the weight distribution and friction of the assembly,
so that no tilt mechanism or auxiliary tool is required.
The operator then removes the thin upper holder bracket
($5$--$10$\,s), the only manual intervention in the
pipeline.

\begin{figure}[!t]
\centering
\begin{subfigure}[t]{0.32\linewidth}
  \centering
  \includegraphics[width=\linewidth,height=0.75\linewidth,keepaspectratio]{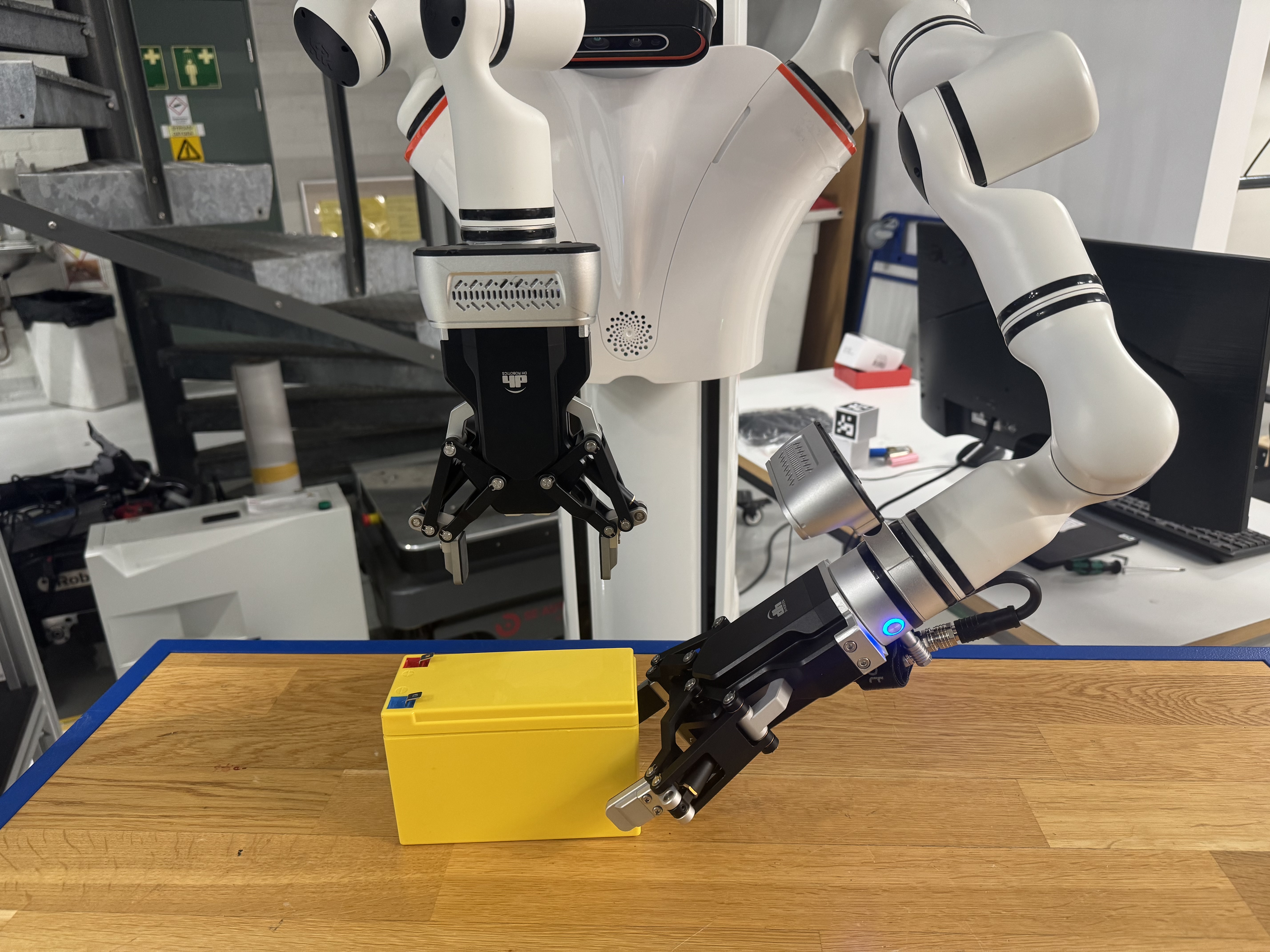}
  \caption{Right-arm lid detection.}
\end{subfigure}\hfill
\begin{subfigure}[t]{0.32\linewidth}
  \centering
  \includegraphics[width=\linewidth,height=0.75\linewidth,keepaspectratio]{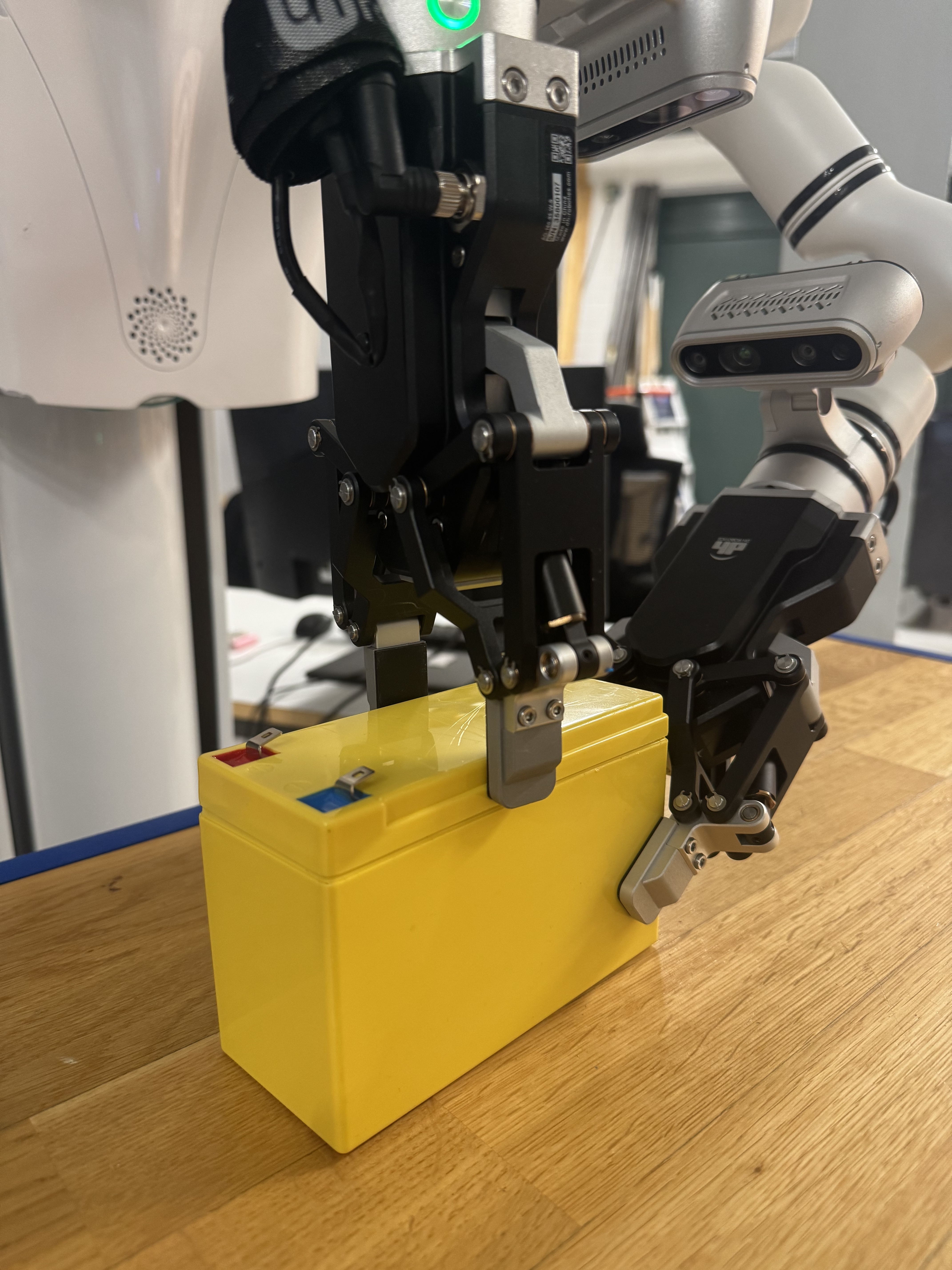}
  \caption{Yaw-aligned descent and grasp.}
\end{subfigure}\hfill
\begin{subfigure}[t]{0.32\linewidth}
  \centering
  \includegraphics[width=\linewidth,height=0.75\linewidth,keepaspectratio]{figures/stage2_lid_removed.jpeg}
  \caption{Lid removed; 21 cells exposed.}
\end{subfigure}\\[0.2cm]
\begin{subfigure}[t]{0.48\linewidth}
  \centering
  \includegraphics[width=\linewidth]{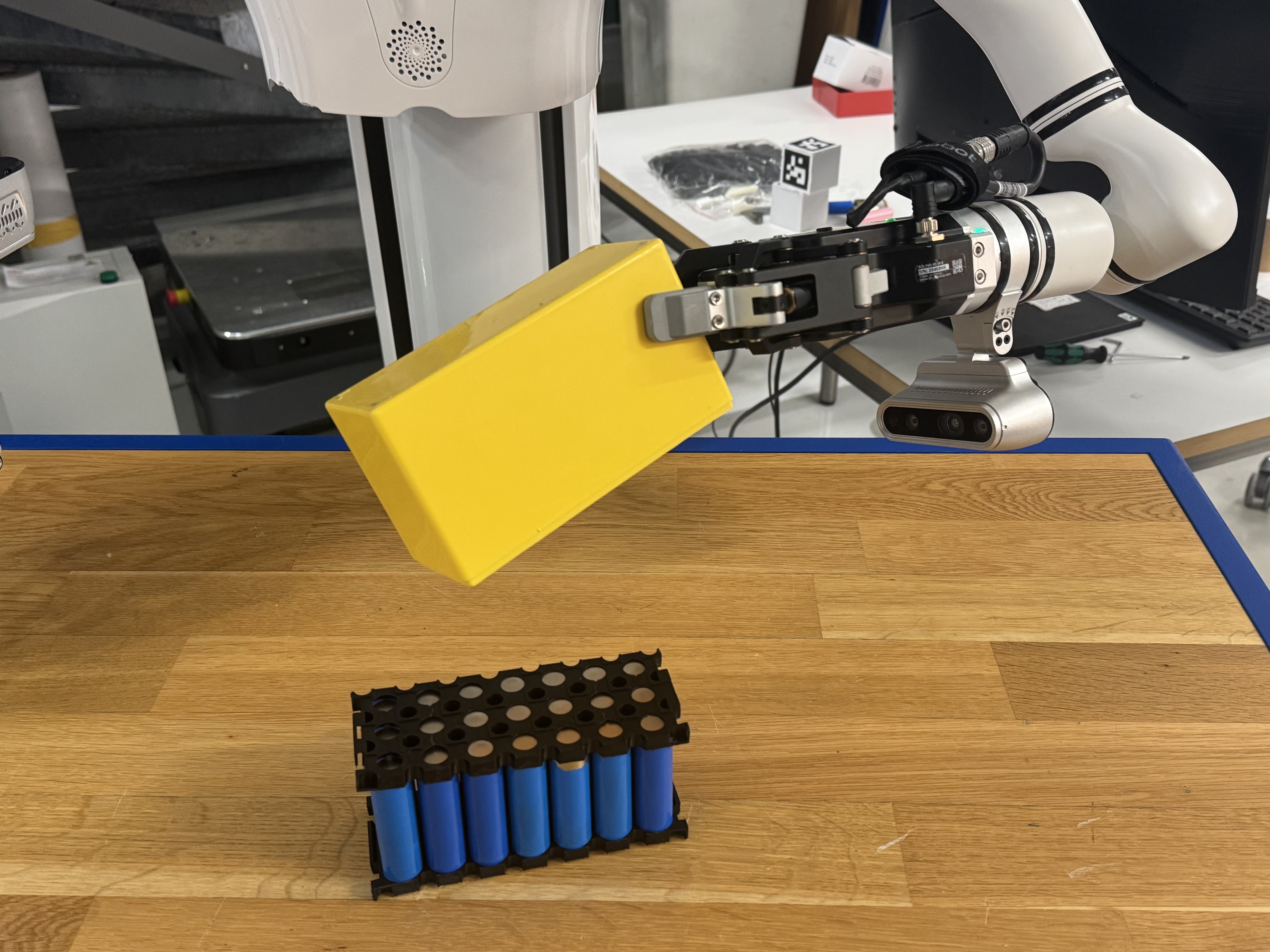}
  \caption{Flip-and-tilt cell dump.}
\end{subfigure}\hfill
\begin{subfigure}[t]{0.48\linewidth}
  \centering
  \includegraphics[width=\linewidth]{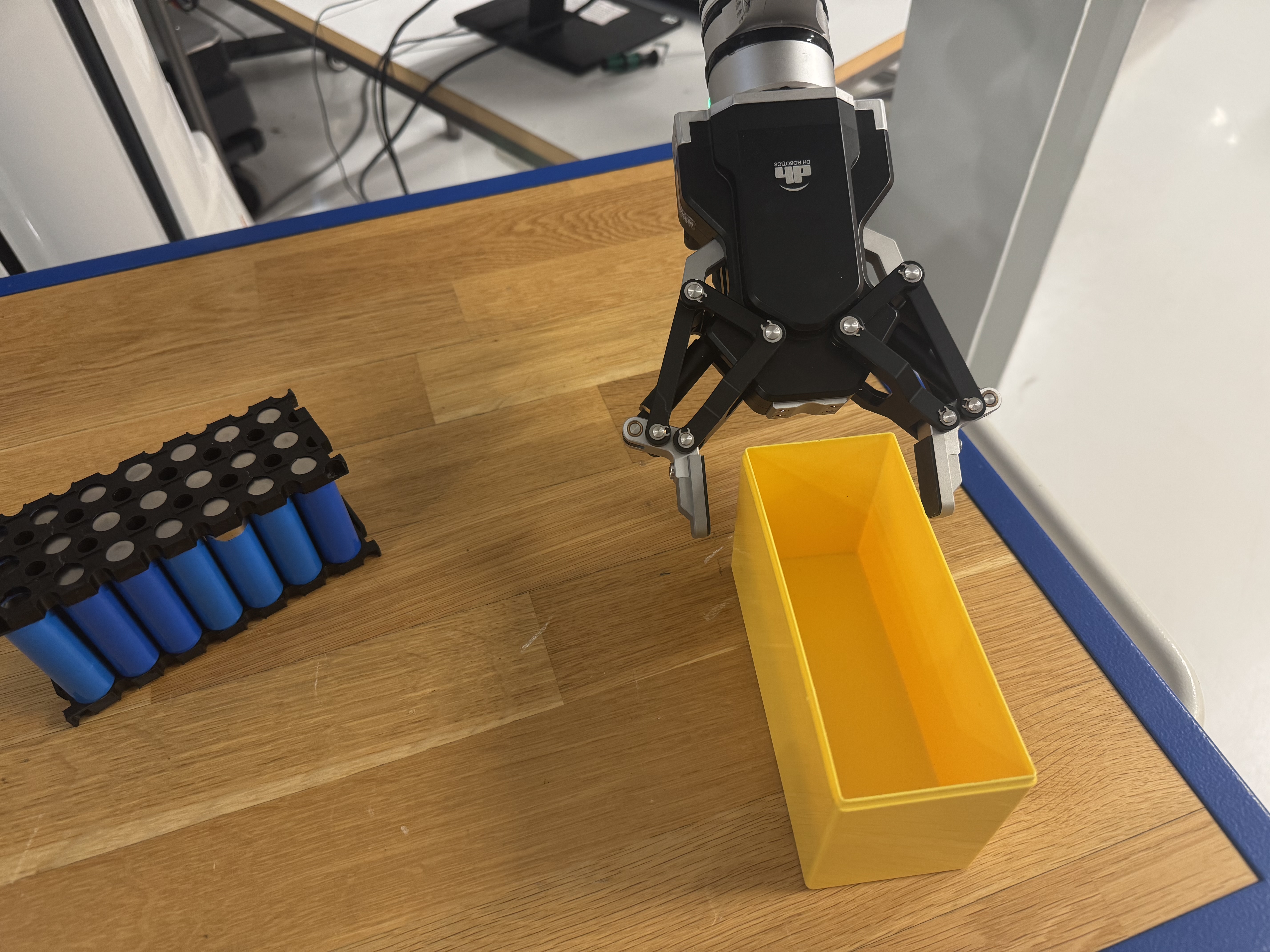}
  \caption{Cell assembly on the workspace.}
\end{subfigure}
\caption{Stage~2: lid detection, lid removal, and cell dump.}
\label{fig:stage2}
\end{figure}

\subsection{Stage 3: Cell assembly relocation}
\label{sec:stage3}

After the manual bracket-removal step, the remaining cell assembly
(the lower holder together with its mounted cells) lies on the
workspace surface in an arbitrary pose.
A new grasp candidate is recovered with
Algorithm~\ref{alg:graspselect}, this time using a workspace box
$\mathcal{W}_3$ tuned to the post-dump region.
The left arm then executes the same approach, grasp, lift, and
transport sequence as in Stage~1, but with reduced grip force.
Full force would squeeze the now-unsupported holder and risk
shifting or detaching individual cells.

The stage concludes with an incremental press-down that seats the
holder against the workspace surface,
\begin{equation}
z_\text{EE}(k+1)=z_\text{EE}(k)-\delta_z,\qquad
k=0,1,\ldots,\lfloor D_\text{max}/\delta_z\rfloor,
\label{eq:pressdown}
\end{equation}
with $\delta_z=5$\,mm and $D_\text{max}=80$\,mm.
The step size is small enough for the arm's stall-detection mechanism
to register holder contact before the holder can deform, and the
maximum descent exceeds the worst-case transport height error while
remaining safely below the depth at which the holder would fracture.

\subsection{Stage 4: Sequential cell extraction}
\label{sec:stage4}

Stage~4 starts with multi-frame HoughCircles cell localisation.
The right wrist camera moves to the overhead observation pose and
runs a five-step localisation per frame.
First, depth gating removes background, worktable, and holder walls
by retaining only depths inside the expected cell-top range.
Second, CLAHE contrast enhancement sharpens edge contrast on the
cell-top surfaces.
Third, a Gaussian blur attenuates high-frequency noise prior to
circle detection.
Fourth, OpenCV's \texttt{HOUGH\_GRADIENT} detector searches for
circles whose radius lies inside bounds derived analytically from
the cell radius $r=9$\,mm and the measured scene depth $d$,
\begin{equation}
r_\text{expected}=\tfrac{r}{d}\,f_x,\quad
r_\text{min}=0.6\,r_\text{expected},\;
r_\text{max}=1.5\,r_\text{expected}.
\label{eq:hough_radius}
\end{equation}
Fifth, a final depth filter discards detections whose centre depth
falls outside the gated range.
For each accepted centre $(u_i,v_i)$ with median patch depth $d_i$,
the 3-D cell position in the base frame is recovered through
Eq.~\eqref{eq:projection}.

Per-frame depth noise is reduced by accumulating $N_\text{avg}=100$
frames and averaging,
\begin{equation}
\hat{\mathbf{c}}_i=\tfrac{1}{|\mathcal{F}_i|}
  \sum_{t\in\mathcal{F}_i}\mathbf{c}_i^{(t)},
\label{eq:multiframe}
\end{equation}
where $\mathcal{F}_i$ collects the frames in which cell~$i$ was
detected. Frames whose total detection count deviates from the
running median are excluded.
This averaging step is the single most important determinant of
extraction accuracy and is examined in detail in
Section~\ref{sec:ablation}.

The extraction order is then planned and a dual-arm support transfer
is scheduled mid-stage.
Cells are sorted in ascending $y_{\mathcal{B}}$, so that the
active arm always reaches the most accessible remaining cell first,
\begin{equation}
\sigma^*_i=
  \underset{j\,\notin\,\{\sigma^*_1,\ldots,\sigma^*_{i-1}\}}
          {\arg\min}\;(\mathbf{c}_j)_y.
\label{eq:order}
\end{equation}
The right arm executes Algorithm~\ref{alg:extract} (Phase~B) until
its workspace boundary is reached.
A support transfer then occurs. The right arm clamps the holder
base at 80\,\% grip force, the left arm releases and returns to home,
and the same algorithm runs again from the opposite side (Phase~C).
The transfer effectively doubles the workspace covered by a single
arm and eliminates the need for an external clamp.
For the final cell, the dual-arm group plans a coordinated
simultaneous release in which the left arm transfers the cell to the
bin while the right arm releases the holder.

\begin{algorithm}[H]
\DontPrintSemicolon
\caption{Cell extraction loop (Phase~B or Phase~C).}
\label{alg:extract}
\KwIn{Wrist observation $\mathcal{O}_\text{wrist}$;
      arm $a\in\{\text{left},\text{right}\}$;
      output bin $\mathcal{D}_a$}
\KwOut{Cells in $\mathcal{D}_a$; remaining unextracted set}
Move arm $a$ to overhead observation pose;\;
$\hat{\mathcal{C}}\leftarrow\textsc{HoughAvg}(\mathcal{O}_\text{wrist},N_\text{avg})$
  \tcp*{Eq.~\eqref{eq:multiframe}}
Sort $\hat{\mathcal{C}}$ by ascending $y_{\mathcal{B}}$;\;
\ForEach{cell $\hat{\mathbf{c}}_i$ (sorted)}{
  Move arm $a$ to pre-grasp pose (table collision active);\;
  Open gripper to extraction width; deactivate table collision;\;
  Cartesian approach to $\hat{\mathbf{c}}_i+\boldsymbol{\delta}_\text{offset}$;\;
  Close gripper (100\,\%); Cartesian vertical lift;\;
  Transport to $\mathcal{D}_a$; open gripper;\;
  \lIf{arm $a$ at workspace limit}{\textbf{break}}
}
\Return $\hat{\mathcal{C}}\setminus\{\text{extracted}\}$;\;
\end{algorithm}

\begin{figure}[!t]
\centering
\begin{subfigure}[t]{0.48\linewidth}
  \centering
  \includegraphics[width=\linewidth]{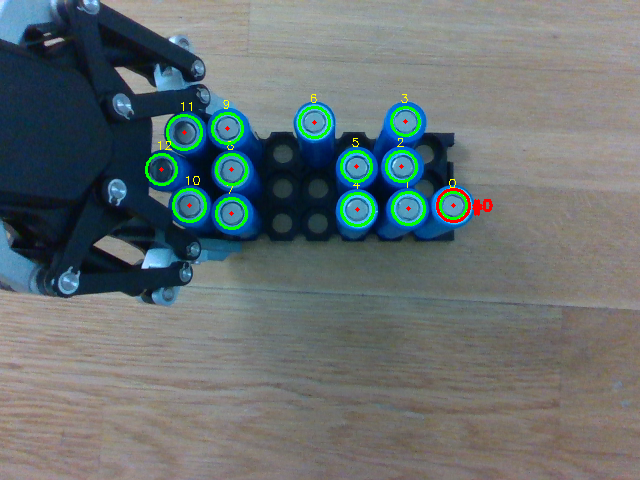}
  \caption{Right wrist (Phase~B): detection on partially extracted assembly.}
\end{subfigure}\hfill
\begin{subfigure}[t]{0.48\linewidth}
  \centering
  \includegraphics[width=\linewidth]{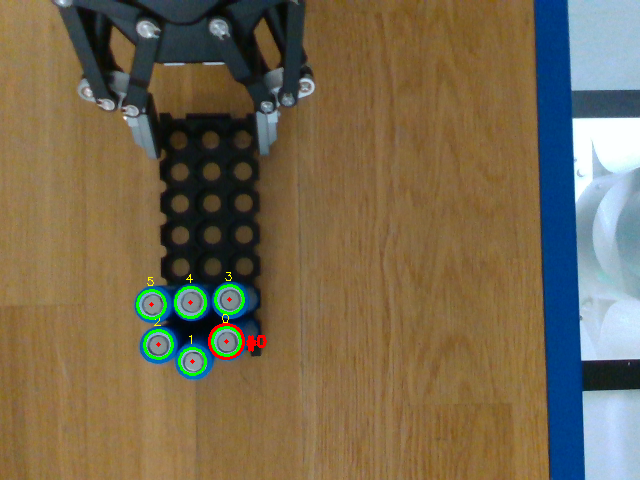}
  \caption{Left wrist (Phase~C): detection on remaining cells.}
\end{subfigure}
\caption{Multi-frame averaged HoughCircles detection. Green circles mark
detected cell tops, and yellow numbers indicate the planned extraction order
(near-to-far in $y_{\mathcal{B}}$).}
\label{fig:cell_detection}
\end{figure}

\begin{figure}[!t]
\centering
\begin{subfigure}[t]{0.32\linewidth}
  \centering
  \includegraphics[width=\linewidth,height=0.75\linewidth,keepaspectratio]{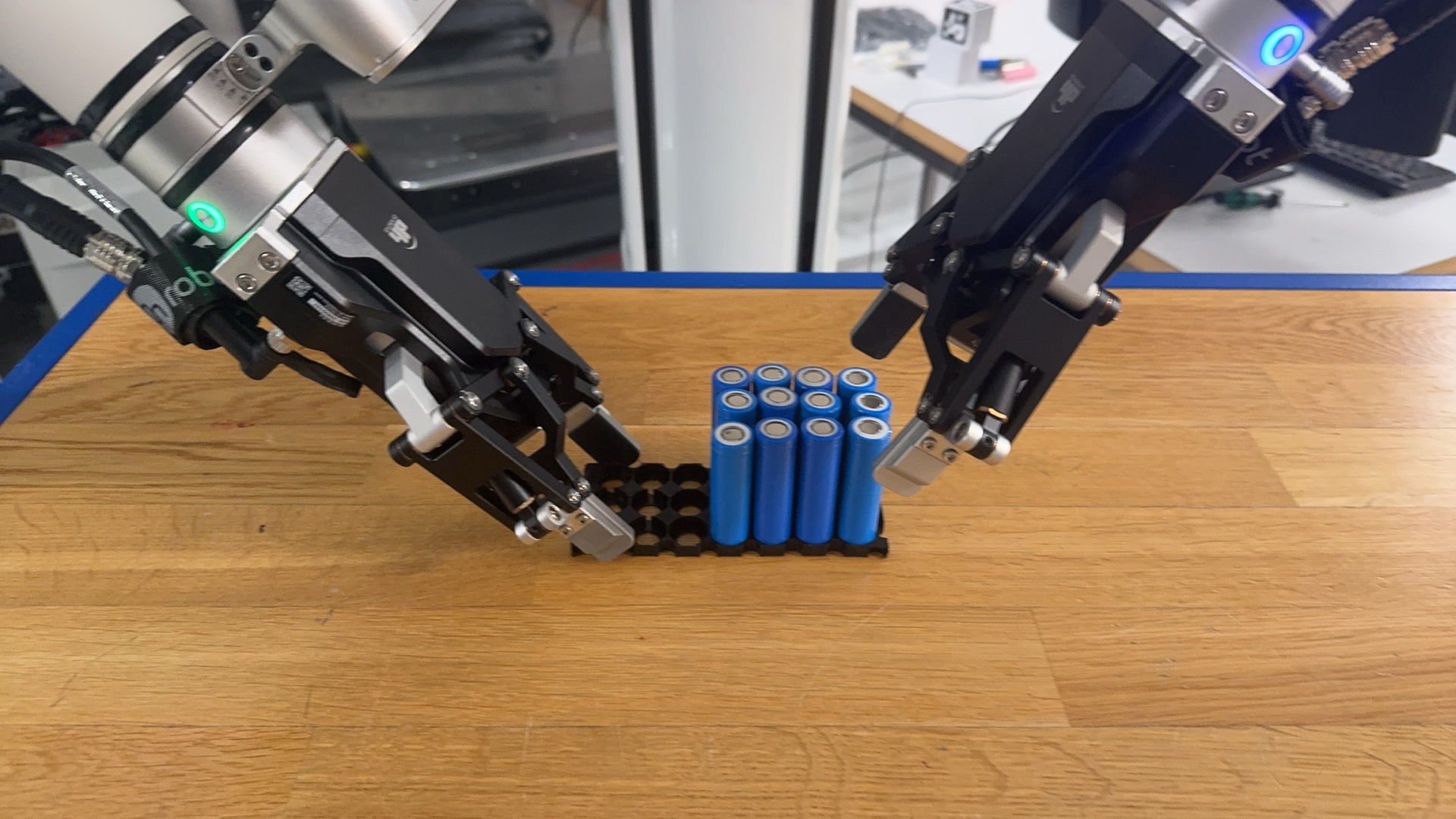}
  \caption{Support transfer: right arm clamps holder.}
\end{subfigure}\hfill
\begin{subfigure}[t]{0.32\linewidth}
  \centering
  \includegraphics[width=\linewidth,height=0.75\linewidth,keepaspectratio]{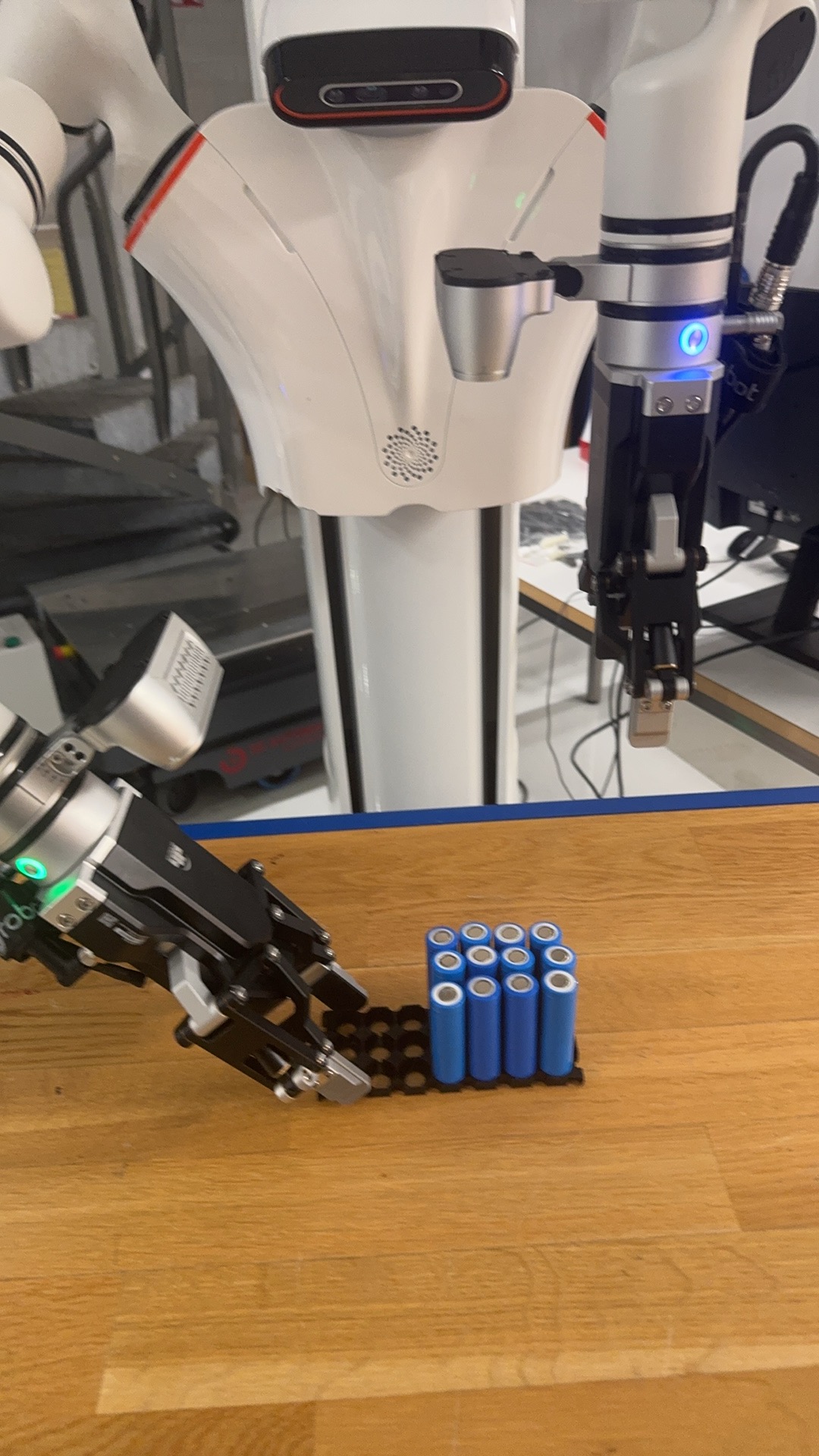}
  \caption{Phase~C: left wrist re-detects remaining cells.}
\end{subfigure}\hfill
\begin{subfigure}[t]{0.32\linewidth}
  \centering
  \includegraphics[width=\linewidth,height=0.75\linewidth,keepaspectratio]{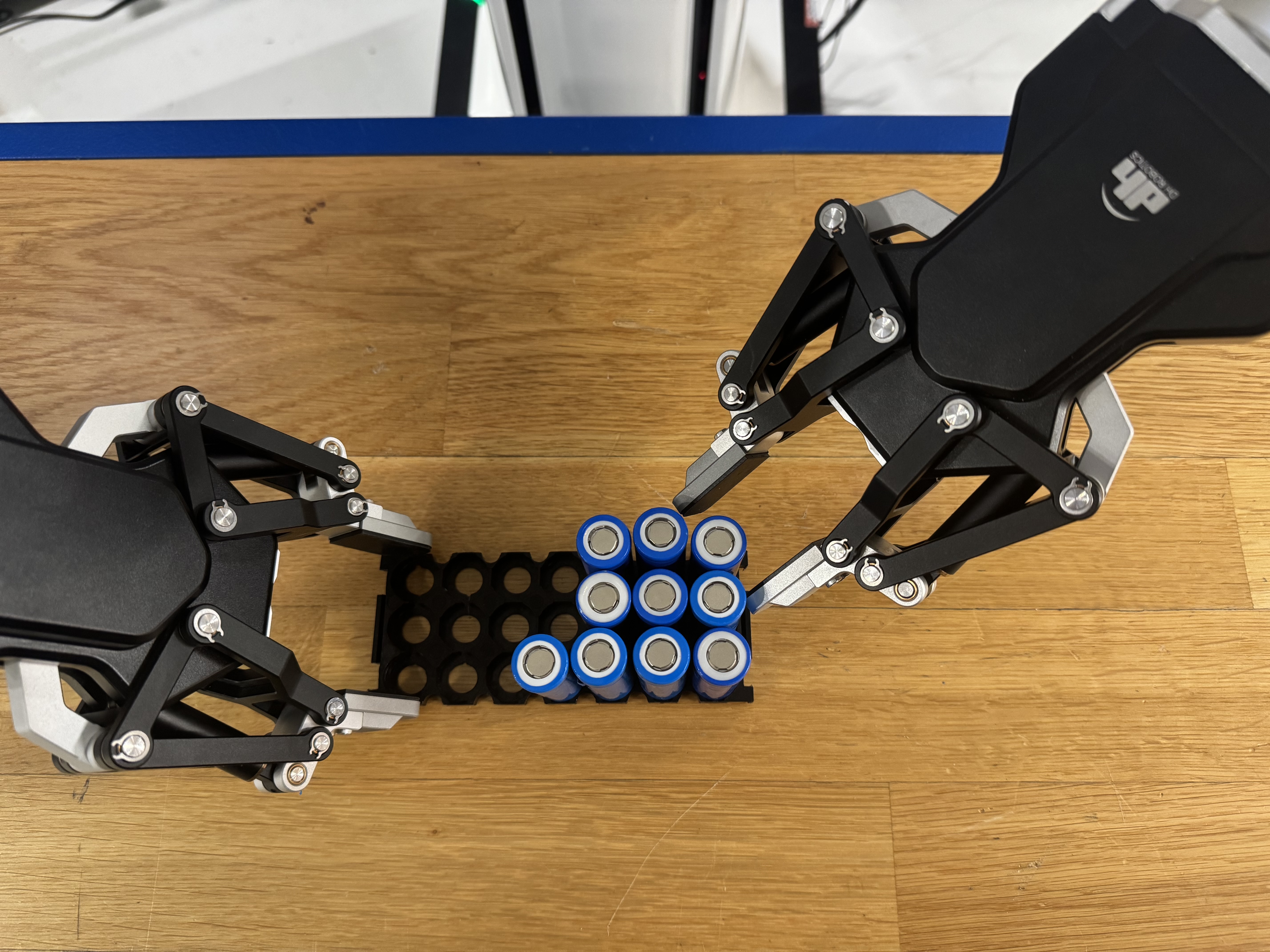}
  \caption{Final dual-arm coordinated release.}
\end{subfigure}
\caption{Stage~4: support transfer between arms and Phase~C extraction.}
\label{fig:stage4_transfer}
\end{figure}

Lightweight error-recovery logic is built into each stage.
If GraspNet returns no valid candidate, the system captures a fresh
frame and retries up to three times before requesting manual
intervention.
If the Cartesian feasibility fraction $f$ drops below
$f_\text{accept}=0.8$, the system falls back to joint-space planning
to the same target, sacrificing straight-line accuracy for
robustness.
HoughCircles outliers are absorbed by the multi-frame averaging in
Eq.~\eqref{eq:multiframe}.
After each vertical lift in Stage~4, a gripper-closure-distance
check determines whether a cell has actually been grasped. A zero
closure distance indicates a missed grasp, in which case the cell is
marked as undetected and skipped.

\section{Experiments}
\label{sec:experiments}

\subsection{Setup and protocol}

All experiments were carried out on a dual-arm platform realising
the architecture of Section~\ref{sec:hw}.
The two manipulators are Realman~RM75 7-DoF arms (5\,kg payload,
610\,mm reach, 0.05\,mm repeatability) mounted on a shared base,
each equipped with a DH~Robotics~AG-160-95 parallel-jaw electric
gripper (95\,mm stroke, 45--160\,N grip force, 0.05\,mm
repeatability, 0.9\,s open/close time).
Three Intel~RealSense~D435 RGB-D cameras (one head-mounted and one
on each wrist) stream at $640\times 480$\,px and 15--30\,fps, with
colour-to-depth alignment enabled through \texttt{rs.align}.
In the present implementation, the left wrist camera handles
lateral correction during pack acquisition and cell localisation in
Phase~C of Stage~4, while the right wrist camera handles lid
detection in Stage~2 and cell localisation in Phase~B of Stage~4.
The compute unit runs Ubuntu~20.04 with ROS~Noetic and MoveIt~1.
Each camera is served by a dedicated daemon thread that maintains
a lock-protected latest-frame buffer so that perception latency does
not block the main planning thread.
The work table measures approximately $1.2\times 0.8$\,m, with
constant laboratory illumination free of strong reflections.
A single 18650 battery-pack mock-up was used throughout. Its external
dimensions and internal cell layout match a commercial 18650 module
(Table~\ref{tab:battery_specs}), and the pack was reassembled
(cells reinserted, lid re-attached) between runs.

For safety, the mock-up is intentionally non-functional. Its geometry
matches a commercial product, but the cells store no electrical
energy and the terminals are not live, which eliminates
thermal-runaway and electrical hazards during robotic handling.
Voltage and temperature monitoring would be required for industrial
deployment on packs that retain residual
charge~\cite{harper2019recycling}, and the present platform does
not include this safety layer.

The trial protocol was the same across all evaluations.
At the start of each trial, the pack was placed within a
$\pm 5$\,cm region around a nominal location reachable by the left
arm, with the yaw randomly varied within $\pm 30^{\circ}$ of the
nominal orientation.
Placements outside this range would require a manipulator with
greater payload and reach and are left to future work.
Following standard practice in robotic manipulation
research~\cite{fang2023anygrasp,kay2022robotic}, 10~trials per
condition were performed for the component evaluations and
10~end-to-end trials for the full pipeline.
A trial is counted as a success if the pipeline completes without
unplanned manual intervention. The designated manual bracket-removal
step between Stages~2 and~3 is not counted.
Timings are measured from the first GraspNet inference call (or
HoughCircles loop start) to the final gripper release.
The $5$--$10$\,s manual bracket-removal and pack-return time
is excluded from all reported timings.

\begin{table}[!t]
\centering
\caption{Battery-pack mock-up specifications.}
\label{tab:battery_specs}
\small
\begin{tabular}{lll}
\toprule
\textbf{Parameter} & \textbf{Value} & \textbf{Notes} \\
\midrule
Enclosure length & 150\,mm & long axis \\
Enclosure width  & 64\,mm  & short axis \\
Enclosure height & 96\,mm  & \\
Lid thickness    & 15\,mm  & snap-fit, no clasps \\
Cell type        & 18650   & cylindrical \\
Cell diameter    & 18\,mm  & \\
Cell height      & 65\,mm  & \\
Cell count       & 21      & 3 rows $\times$ 7 columns \\
Cell spacing     & 1\,mm   & edge-to-edge gap \\
\bottomrule
\end{tabular}
\end{table}

Table~\ref{tab:planning_params} lists the MoveIt parameters used
throughout, and Table~\ref{tab:stage_params} consolidates the key
stage-specific parameters.

\begin{table}[!t]
\centering
\caption{MoveIt planning parameters used in all stages.}
\label{tab:planning_params}
\small
\begin{tabular}{lll}
\toprule
\textbf{Parameter} & \textbf{Value} & \textbf{Description} \\
\midrule
Planning time       & 15\,s  & Maximum time per planning query \\
Planning attempts   & 5      & Parallel attempts \\
Position tolerance  & 10\,mm & Goal position tolerance \\
Orientation tolerance & 0.10\,rad & Goal orientation tolerance \\
Velocity scaling    & 1.0    & Fraction of max joint velocity \\
Acceleration scaling & 1.0   & Fraction of max joint acceleration \\
Nearest-IK $d_\text{stop}$ & 0.5\,rad & Early-stop threshold \\
Cartesian step $\delta$ (precision)  & 5\,mm  & Near object \\
Cartesian step $\delta$ (transport)  & 10\,mm & Long moves \\
Cartesian acceptance $f_\text{accept}$ & 0.8 & Fallback threshold \\
\bottomrule
\end{tabular}
\end{table}

\begin{table}[!t]
\centering
\footnotesize
\caption{Key stage-specific parameters and their rationale.}
\label{tab:stage_params}
\begin{tabularx}{\linewidth}{p{3.5cm}>{\raggedright\arraybackslash}p{3cm}X}
\toprule
\textbf{Parameter} & \textbf{Value} & \textbf{Rationale} \\
\midrule
\multicolumn{3}{l}{\textit{Stage~1: Pack acquisition}}\\
GraspNet sample count $N_\text{pts}$ & 20\,000 & Default of GraspNet-Baseline. \\
Front-surface bias $d_\text{front}$ & 50\,mm & Targets the accessible front face. \\
Approach threshold $\epsilon_1$ & 0.3 & Allows approximately $17^{\circ}$ deviation from horizontal. \\
Score threshold $\tau$ & 0.1 & Lowest value rejecting boundary spurious proposals. \\
PCA radius $r_\text{pca}$ & 100\,mm & Covers the 64\,mm short axis with margin. \\
Wrist offset $\delta_u$ & 108\,px & One-time calibration; consistent with camera-to-TCP offset. \\
Lateral deadband & $\pm 10$\,mm & Suppresses depth-sensor noise. \\
\midrule
\multicolumn{3}{l}{\textit{Stage~2: Lid removal \& cell dump}}\\
Lid frames $N_\text{avg}$ & 30 & Brings $z$-std below 1\,mm. \\
Depth gate $[d_\text{min},d_\text{max}]$ & $[0.214,0.232]$\,m & Matches wrist-to-lid standoff. \\
HSV gate $\mathcal{H}_\text{lid}$ & $H\!\in\![26^{\circ},83^{\circ}]$, $S\!\in\![18,86]$, $V\!\in\![95,255]$ & Isolates the yellow lid under lab lighting. \\
Outlier gate $\delta_z^\text{gate}$ & 2\,mm & Rejects frames deviating from running median. \\
Cell-dump pitch & $-30^{\circ}$ & Shifts contact centroid to proximal gripper face. \\
Force reduction & 10\,\% & Lets cells slide out while shell stays held. \\
\midrule
\multicolumn{3}{l}{\textit{Stage~3: Assembly relocation}}\\
Pre-grasp distance & 130\,mm & Clearance above 96\,mm enclosure with 30\,mm margin. \\
Press-down step $\delta_z$ & 5\,mm & Triggers stall detection without damaging holder. \\
Max depth $D_\text{max}$ & 80\,mm & Below the depth that would fracture the holder. \\
\midrule
\multicolumn{3}{l}{\textit{Stage~4: Cell extraction}}\\
Averaging frames $N_\text{avg}$ & 100 & $1/\sqrt{100}=10\times$ noise reduction. \\
Gripper opening & 30\,mm & 6\,mm clearance per side around 18\,mm cell. \\
Wall offset $\boldsymbol{\delta}_\text{offset}$ & $(-18,+18,-25)$\,mm & Compensates slanted holder walls. \\
Support-transfer force & 80\,\% & Stabilises holder without crushing it. \\
Vertical lift & 80\,mm & Clears holder walls; used for slip detection. \\
\bottomrule
\end{tabularx}
\end{table}

\subsection{Per-stage results}
\label{sec:results}

Tables~\ref{tab:stage1_grasp}--\ref{tab:cell_localise} summarise the
per-stage results.
With the wrist-camera lateral correction enabled, Stage~1 reaches
10/10 success at a mean cycle time of 28.0\,s
(Table~\ref{tab:stage1_grasp}). The rare failures observed without
the correction occur at extreme yaw ($|\psi|>25^{\circ}$ from
nominal), where the PCA alignment becomes unreliable.
Stage~2 succeeded in every sub-step on every trial, including lid
detection, lid removal, and cell dump
(Table~\ref{tab:stage2_grasp}), which confirms that the
fused depth and colour signal is sufficient for a coloured snap-fit lid and
that the flip-and-tilt motion reliably overcomes the static friction
between the cell assembly and its case.
Stage~3 was likewise uniformly successful
(Table~\ref{tab:stage3_reloc}), with the incremental press-down
reaching its depth limit in all trials.
In Stage~4, all 21~cells were detected in every trial and on
average $20.7/21$ cells were extracted
(Table~\ref{tab:cell_localise}). The rare misses correspond to
extraction slips during the vertical lift rather than to detection
failures.

\begin{table}[!t]
\centering
\caption{Stage~1 pack-acquisition evaluation
($N=10$ trials per condition).}
\label{tab:stage1_grasp}
\small
\begin{tabular}{lcc}
\toprule
\textbf{Condition} & \textbf{Success} & \textbf{Mean time (s)} \\
\midrule
With lateral correction    & 10/10 & 28.0 \\
Without lateral correction & 7/10  & 26.6 \\
\bottomrule
\end{tabular}
\end{table}

\begin{table}[!t]
\centering
\caption{Stage~2 lid removal \& cell dump ($N=10$).}
\label{tab:stage2_grasp}
\small
\begin{tabular}{lcc}
\toprule
\textbf{Sub-step} & \textbf{Success} & \textbf{Mean time (s)} \\
\midrule
Lid detection (30-frame)  & 10/10 & 5.5 \\
Lid removal               & 10/10 & 16.5 \\
Cell dump (flip+tilt)     & 10/10 & 17.7 \\
\midrule
\textbf{Stage~2 overall}  & 10/10 & 39.7 \\
\bottomrule
\end{tabular}
\end{table}

\begin{table}[!t]
\centering
\caption{Stage~3 assembly relocation ($N=10$).}
\label{tab:stage3_reloc}
\small
\begin{tabular}{lcc}
\toprule
\textbf{Metric} & \textbf{Value} & \textbf{Notes} \\
\midrule
Grasp + relocation success & 10/10 & \\
Press-down completed       & 10/10 & reached depth limit \\
Mean Stage~3 time          & 20.9\,s & \\
\bottomrule
\end{tabular}
\end{table}

\begin{table}[!t]
\centering
\caption{Stage~4 cell localisation and extraction ($N=10$ replicates,
100-frame averaging per detection).}
\label{tab:cell_localise}
\small
\begin{tabular}{lcc}
\toprule
\textbf{Metric} & \textbf{Value} & \textbf{Notes}\\
\midrule
Full-count detection (21/21) & 10/10  & \\
Mean extraction success      & 20.7/21 & cells per trial \\
\bottomrule
\end{tabular}
\end{table}

\subsection{End-to-end results}
\label{sec:e2e}

Table~\ref{tab:e2e} summarises the ten end-to-end trials.
The pipeline reached full success (all 21~cells extracted) in 8 out
of 10 trials, extracting on average $20.6/21$~cells per trial with a
mean cycle time of $6.0$\,minutes (excluding the manual
bracket-removal step).
Stage~4 dominates the cycle at approximately 75\,\% of the total
time, driven by the 100-frame averaging window and the cell-by-cell
extraction loop. Both are explicit accuracy and throughput trade-offs
that the system designer can retune for higher throughput at the cost
of reduced robustness (Section~\ref{sec:ablation}).
The observed failure modes (Table~\ref{tab:failure_modes}) both occur
in Stage~4. The first is a single incomplete cell detection caused by
corner cells occluded by the holder, and the second is a single
extraction slip during the vertical lift.
No Stage~1--3 failures were observed in the ten end-to-end trials.

\begin{table}[!t]
\centering
\caption{End-to-end disassembly results ($N=10$).
Manual bracket-removal time is excluded from all timings.}
\label{tab:e2e}
\small
\begin{tabular}{lcc}
\toprule
\textbf{Metric} & \textbf{Value} & \textbf{Notes} \\
\midrule
Full success (21/21 cells)      & 8/10    & \\
Mean cells extracted per trial  & 20.6/21 & \\
\midrule
Mean Stage~1 time               & 28.0\,s & pack acquisition \\
Mean Stage~2 time               & 39.7\,s & lid removal \& cell dump \\
Mean Stage~3 time               & 20.9\,s & assembly relocation \\
Mean Stage~4 time               & 271.2\,s& all cell extractions \\
\textbf{Mean total time}        & 6.0\,min& excl.\ manual step \\
\bottomrule
\end{tabular}
\end{table}

\begin{table}[!t]
\centering
\caption{Failure mode summary across all end-to-end trials.}
\label{tab:failure_modes}
\footnotesize
\begin{tabularx}{\linewidth}{p{1.0cm}p{3.2cm}cX}
\toprule
\textbf{Stage} & \textbf{Failure mode} & \textbf{Count} & \textbf{Root cause} \\
\midrule
1 & Grasp miss (closed on air) & 0 & Lateral offset above dead-band or extreme yaw \\
1 & GraspNet: no valid candidate & 0 & Point cloud too sparse / occluded \\
2 & Lid slip during removal & 0 & Grasp centre near lid edge \\
2 & Incomplete cell dump   & 0 & Cells stuck to enclosure wall \\
4 & Incomplete cell detection & 1 & Corner cells occluded by holder \\
4 & Cell extraction slip   & 1 & Cell fell during lift \\
\bottomrule
\end{tabularx}
\end{table}

\subsection{Ablation studies}
\label{sec:ablation}

Two targeted ablations isolate the contribution of the perception
choices to overall reliability.

The first ablation isolates the effect of the wrist-camera lateral
correction in Stage~1.
Table~\ref{tab:stage1_grasp} compares the stage with and without the
single-shot correction described in Section~\ref{sec:stage1}.
Disabling the correction reduces the Stage~1 success rate from
$10/10$ to $7/10$. With no compensation, the gripper systematically
misses the pack whenever the residual head-camera lateral error
exceeds the parallel-jaw capture range, confirming the analysis in
Section~\ref{sec:stage1}.
The cost of enabling the correction (a single depth frame and a
re-execution of the pre-grasp approach) is negligible relative to
the 30-percentage-point improvement in success rate.

The second ablation examines multi-frame averaging in Stage~4.
Ten repeated measurements were performed on the same fixed cell
layout, with and without the 100-frame averaging from
Eq.~\eqref{eq:multiframe}.
Under the assumption of independent per-frame noise, averaging
$N_\text{avg}=100$ frames should reduce position noise by a factor of
$1/\sqrt{100}=10$.
The empirical root-mean-square (RMS) standard deviation drops from
$18.7$\,mm under the single-frame condition to $2.4$\,mm under
100-frame averaging
(Table~\ref{tab:ablation_mfa}), a factor of approximately $7.8$ that
is close to the theoretical upper bound and brings the residual error
well within the $\pm 6$\,mm per-side clearance around an 18\,mm cell
inside the 30\,mm gripper opening.
Increasing $N_\text{avg}$ beyond 100 yields diminishing accuracy
gains while linearly increasing the per-cell sensing time, so 100
frames is chosen as the operating point for the end-to-end
evaluation.

\begin{table}[!t]
\centering
\caption{Ablation: multi-frame averaging effect on cell-position
repeatability ($N=10$ repeated measurements of the same layout).}
\label{tab:ablation_mfa}
\small
\begin{tabular}{lcc}
\toprule
\textbf{Condition} & $\sigma_\text{RMS}$ (mm) & Full det. \\
\midrule
Single frame ($N_\text{avg}=1$)     & 18.7 & 6/10 \\
100-frame avg ($N_\text{avg}=100$)  & 2.4  & 9/10 \\
\bottomrule
\end{tabular}
\end{table}

\subsection{Comparison with related work}
\label{sec:comparison}

Table~\ref{tab:comparison} positions the present pipeline against
representative robotic battery-disassembly systems.
The works differ in decomposition level (pack, module, or cell),
hardware and software configuration, and operating environment
(simulation or real-world), so the comparison is meant for context
rather than as a strict benchmark.
To the best of our knowledge, ours is the first system to use a dual-arm
robot and parallel-jaw grippers to perform 18650 cell-level extraction
without fixed fixtures, on a pack at an unknown initial pose, while
combining data-driven 6-DoF grasp estimation with rule-based perception
and discrete visual correction.

\begin{table}[!t]
\centering
\caption{Comparison with related robotic battery-disassembly systems.
Level: P\,=\,pack, M\,=\,module, C\,=\,cell.
Cycle times marked ``n/r'' were not reported.}
\label{tab:comparison}
\small
\begin{tabularx}{\linewidth}{@{}X c c c l l@{}}
\toprule
\textbf{Work} & \textbf{Level} & \textbf{Pose unc.\textsuperscript{a}} &
\textbf{Arms} & \textbf{Sensors} & \textbf{Cycle} \\
\midrule
Kay et al.~\cite{kay2022robotic}
  & M & No & 1 & F/T sensor & n/r \\
\quad\textit{\footnotesize Force-controlled extraction} & & & & & \\
Contreras et al.~\cite{contreras2024multirobot}
  & M & Partial & 4 (sim) & Cam end-eff. & n/r \\
\quad\textit{\footnotesize RRT multi-robot planning} & & & & & \\
Qu et al.~\cite{qu2024robotic}
  & M & Partial & 4 & 2 cameras & n/r \\
\quad\textit{\footnotesize Two-step visual localisation} & & & & & \\
Liang et al.~\cite{liang2025experimental}
  & M & No & 1 & Camera + F/T & 32.7\,min\textsuperscript{b} \\
\quad\textit{\footnotesize Structured disassembly} & & & & & \\
\textbf{Ours}
  & \textbf{C} & \textbf{Yes} & \textbf{2} & 3 RGB-D & \textbf{6.0\,min} \\
\quad\textit{\footnotesize GraspNet + wrist-cam correction + HoughCircles} & & & & & \\
\bottomrule
\end{tabularx}\\[2pt]
\raggedright\footnotesize
\textsuperscript{a}\,System handles unknown initial pose without fixtures.\quad
\textsuperscript{b}\,32.7\,min per module (8 prismatic cells); direct
comparison with the present cylindrical-cell pipeline is
approximate.
\end{table}

\section{Discussion}
\label{sec:discussion}

Three findings emerge from the per-stage and end-to-end results.
First, the wrist-camera lateral correction is essential at Stage~1.
Without it, three of ten trials fail at the initial grasp because
the residual head-camera lateral error exceeds the parallel-jaw
capture range (Table~\ref{tab:stage1_grasp}).
A single depth frame and a re-execution of the pre-grasp approach is
sufficient to absorb this error, suggesting that, on platforms
constrained by per-call planning latency, a look-and-move correction
placed at the right moment recovers most of the practical benefit of
continuous visual servoing.
Second, the 100-frame averaging brings cell-position noise from
$18.7$\,mm down to $2.4$\,mm RMS
(Table~\ref{tab:ablation_mfa}), a factor close to the
$1/\sqrt{N_\text{avg}}$ theoretical bound, which places the residual
error inside the $\pm 6$\,mm per-side clearance around an 18\,mm
cell.
This noise reduction is what makes a classical detector with a known
geometric prior competitive with learned alternatives at this scale.
Third, the flip-and-tilt cell-dump primitive is empirically robust
for this pack. A $+180^{\circ}$ wrist flip combined with a small
additional pitch (approximately $20^{\circ}$) reliably overcomes the
static friction between the cell assembly and its case, eliminating
the need for a dedicated tilting mechanism or auxiliary tool.

The role of GraspNet in the proposed pipeline deserves explicit
comment.
The network is used as a frozen, pre-trained candidate generator
followed by domain-specific filtering (workspace boundaries,
approach angle, front-surface bias, and score threshold), rather
than as a black-box grasp predictor.
The proposal stage benefits from the network's ability to produce
geometrically reasonable 6-DoF candidates on reflective enclosures
without any task-specific training data, while the filtering stage
enforces the predictability and safety constraints required for
industrial execution.
For structured disassembly tasks with known target geometries, this
learn-and-filter strategy currently yields more reliable execution
than either pure learning or pure rule-based perception in
isolation, and the same principle is expected to apply to other
end-of-life processing pipelines whose components are geometrically
constrained but spatially uncertain.

The scope and limitations of the present evaluation also merit
discussion.
The pipeline is designed and validated for one commercial 18650 pack
geometry. Other pack sizes or cell layouts can in principle be
supported through parameter rescaling, but they require revalidation.
All experiments were carried out on a non-functional mock-up, and
deployment on energised packs would require an additional safety
layer with voltage and temperature monitoring.
Beyond this general scope, several specific limitations apply.
Only a single mock-up was used, so inter-sample differences are not
captured, and repeated assembly may gradually weaken the lid-retention
force in ways that systematically affect Stage~2.
The initial pose range was set conservatively to $\pm 5$\,cm and
$\pm 30^{\circ}$ around the nominal placement, matched to the RM75's
610\,mm reach, and performance outside this range is untested.
Filter parameters and gripper opening widths are tuned to the
specific 18650 geometry, while the HSV lid filter and HoughCircles
parameters are tuned to constant laboratory illumination.
Both would have to be revisited for substantially different working
conditions.
Cell extraction is currently open-loop. A slip during the lift
cannot be corrected in real time, and success is only detected
post-hoc through the gripper-distance check.
The upper holder bracket still requires a $5$--$10$\,s manual
removal step.
Finally, the ten end-to-end trials follow common practice in robotic
manipulation research~\cite{kay2022robotic,fang2023anygrasp} but
yield wide confidence intervals. An observed $8/10$ success rate
corresponds to a Wilson 95\,\% confidence interval of approximately
$[49\,\%,96\,\%]$, so the precise numerical success rate should be
interpreted with caution, and larger-scale trials would be needed
for narrower bounds.

\section{Conclusions}
\label{sec:conclusion}

This paper presented an end-to-end dual-arm robotic pipeline that
disassembles an 18650 battery pack from an unknown initial pose
using only parallel-jaw grippers, commodity RGB-D sensing, and a
pre-trained grasp detector, without any external fixture or custom
tooling.
A hybrid learn-and-filter perception stack, consisting of GraspNet
candidates refined by workspace filtering, PCA yaw alignment, and a
single-shot wrist-camera lateral correction, together with a
fused depth and colour lid detector, and a multi-frame HoughCircles cell
localiser, is combined with an in-task dual-arm support transfer to
cover the full $3\times 7$ cell array.
On a Realman~RM75 platform equipped with three RealSense~D435
cameras, the pipeline reached an end-to-end success rate of $8/10$,
a cell-localisation accuracy of $\pm 2.4$\,mm (RMS), and a mean
cycle time of 6.0\,minutes per pack, with ablation studies
confirming the necessity of both the lateral correction and the
multi-frame averaging.
Future work will focus on closed-loop cell extraction with
force sensing, automation of the remaining manual bracket-removal
step, and extension to other pack geometries.

\section*{Acknowledgements}
This work was supported by Centre of Excellence in Production Research (XPRES).

\section*{Declaration of competing interest}
The authors declare no competing interests.

\section*{Declaration of generative AI and AI-assisted technologies in the manuscript preparation process}
During the preparation of this work the author(s) used Anthropic Opus 4.7 in order to help preparing the initial draft of this manuscript,  designing code for experiments, and polishing writing. After using this tool/service, the author(s) reviewed and edited the content as needed and take(s) full responsibility for the content of the published article.
\bibliographystyle{elsarticle-num}
\bibliography{references}

\end{document}